\documentclass[preprint,5p,times,twocolumn, sort&compress]{elsarticle}
\usepackage{graphicx, epstopdf}
\usepackage{amsmath}
\usepackage{amssymb}
\usepackage{caption}
\usepackage{subcaption}
\usepackage{hyperref}
\usepackage[table]{xcolor}

\usepackage{array,multirow,graphicx,makecell}

\graphicspath{ {figs/} }
\usepackage{tikz}

\newcommand{\fig}[1]{Fig.\@~#1}
\newcommand{\tab}[1]{Tab.\@~#1}

\usepackage{eso-pic}
\usepackage{transparent}
 
\AddToShipoutPictureBG{
  \AtPageLowerLeft{%
    \raisebox{20pt}{\makebox[\paperwidth]{\begin{minipage}{21cm}\centering \transparent{0.4}
      \centering Author version of the manuscript accepted for publication: 
      Svarny, P.; Rozlivek, J.; Rustler, L.; Sramek, M.; Deli, \"O.; Zillich, M. \& Hoffmann, M. (2022),
       'Effect of Active and Passive Protective Soft Skins on Collision Forces in Human-robot Collaboration', Robotics and Computer-Integrated Manufacturing.  (C) Elsevier. \url{https://doi.org/10.1016/j.rcim.2022.102363}
    \end{minipage}}}%
  }
}

\begin{document}
\begin{frontmatter}
\title{Effect of Active and Passive Protective Soft Skins on Collision Forces in Human-robot Collaboration}

\author[1]{Petr Svarny} 
\author[1]{Jakub Rozlivek}
\author[1]{Lukas Rustler}
\author[1]{Martin Sramek}
\author[2]{\"Ozg\"ur Deli}
\author[2]{Michael Zillich}
\author[1]{Matej Hoffmann\texorpdfstring{\corref{cor1}}{}}\ead{matej.hoffmann@fel.cvut.cz}

\address[1]{Department of Cybernetics, Faculty of Electrical Engineering, Czech Technical University in Prague, Czech Republic}
\address[2]{Blue Danube Robotics GmbH, Austria}
\cortext[cor1]{Corresponding author}

\begin{abstract}
Soft electronic skins are one of the means to turn a classical industrial manipulator into a collaborative robot. For manipulators that are already fit for physical human-robot collaboration, soft skins can make them even safer. In this work, we study the after impact behavior of two collaborative manipulators (UR10e and KUKA LBR iiwa) and one classical industrial manipulator (KUKA Cybertech), in the presence or absence of an industrial protective skin (AIRSKIN). In addition, we isolate the effects of the passive padding and the active contribution of the sensor to robot reaction. We present a total of 2250 collision measurements and study the impact force, contact duration, clamping force, and impulse. This collected dataset is publicly available. We summarize our results as follows. For transient collisions, the passive skin properties lowered the impact forces by about 40~\%. During quasi-static contact, the effect of skin covers---active or passive---cannot be isolated from the collision detection and reaction by the collaborative robots. Important effects of the stop categories triggered by the active protective skin were found. We systematically compare the different settings and compare the empirically established safe velocities with prescriptions by the ISO/TS 15066. In some cases, up to the quadruple of the ISO/TS 15066 prescribed velocity can comply with the impact force limits and thus be considered safe. We propose an extension of the formulas relating impact force and permissible velocity that take into account the stiffness and compressible thickness of the protective cover, leading to better predictions of the collision forces. At the same time, this work emphasizes the need for in situ measurements as all the factors we studied---presence of active/passive skin, safety stop settings, robot collision reaction, impact direction, and, of course, velocity---have effects on the force evolution after impact.
\end{abstract}

\begin{keyword} 
physical HRI \sep protective skin sensors \sep power and force limiting \sep collaborative robots \sep collision measurements \sep robot collision reaction

\end{keyword}

\end{frontmatter}

\section{Introduction}
\label{sec:intro}
Human-robot collaboration (HRC), i.e., the ``continuous, purposeful interaction associated with potential or accidental physical events''~\cite{Vicentini2020a}, promises the empowerment of the human partner by allowing the human to benefit from the superior robot precision, speed, or power. However, for this purpose, the machine needs to be safe. There is a vivid discussion about the correct ways how to ensure safety in a HRC scenario (see \cite{Gualtieri2021}, \cite{Mansfeld2018}, \cite{Lucci2020}, \cite{Bi2021} or \cite{Gopinath2021} for an example of the evaluation of a HRC scenario). One of these approaches is the use of artificial skins as a supporting component to ensure safe HRC and the investigation of their contribution is the aim of this paper.

Gualtieri et al.~\cite{Gualtieri2021} present a division of safety research into two general areas: \textit{prevention}, i.e., trying to avoid a collision, and \textit{protection}, i.e., mitigating the effect of a collision~\cite{Vicentini2020a}. For prevention, artificial skins can use anticipatory mechanisms like proximity sensors or capacitive sensors that sense also the robot surroundings (e.g., Bosch APAS). Electronic skins may also be used to display the robot states and thus facilitate human-robot interaction~\cite{Tang2019}. However, unless these mechanisms provide a sufficiently long-range detection capability that would allow the robot to react, the main contribution of skins to safety is in protection. Therefore the protective capabilities of safety skins are the focus of this work.

The situation where collisions of an operator with a moving robot are allowed is regulated by the Power and Force Limiting (PFL) regime described in the ISO/TS~15066~\cite{ISOTS15066} (TS~15066 in what follows), which encompasses the case where robot skin serves for protection. The regime deals with scenarios where the human and robot get into contact by limiting the forces and pressures exerted on the human.
The specific limits imposed by the standard are debated \cite{Mansfeld2018}, \cite{Svarny2021}, \cite{Han2018}, \cite{Park2019}. However, we will use them as a baseline from which we draw the corresponding force or velocity limits. 

The protective use of artificial skin is not limited only to its role as passive padding~\cite{Vicentini2020a}. If it is sensorized, it can actively detect and localize contacts and thus partake in the collision detection and isolation phases of the collision pipeline~\cite{Haddadin2017}. 
Multiple variables need to be taken into account with regards to skins and the protection provided by them (see discussion in~\cite{Vicentini2020}). 

The overall material properties of protective skins determine their effects on safety --- both for passive shock absorption and for active collision detection. Tsuji et al.~\cite{Tsuji2020}, for example, report a trade-off whereby greater thickness of the skin, improves the passive properties of the skin for protection but impedes proximity sensing prior to the collision. In this work, we assess the stiffness and force threshold of the skin at different locations on the surface.

Various properties also influence the collision forces, e.g., the robot weight or the human body stiffness. These properties can be accounted for in an analytical model. A well-established contact model is the Hertz contact model. However, even with improved models as \cite{Vemula2017, Vemula2018}, many safety-related aspects of the collision can be missed by a purely analytical model. Especially post-collision reactions can change the resulting exerted forces (see \cite{Haddadin2016}), favoring experimental studies of collision forces, as shown in \cite{Schlotzhauer2019,Svarny2021} where data-driven collision-force-maps are established. The evaluation of collision effects (namely motor currents or forces) can serve by itself, even without a sensing skin, as a safety measure (see for example~\cite{Aivaliotis2019} for a recent approach) and is part of modern collaborative robots. Therefore we take these robot-specific capabilities also into consideration during our experiments.

The goal of this paper is to investigate the contribution of soft, protection-intended skin components on the overall safety of a collaborative robot system through a comprehensive study on multiple robots and with different settings. We provide a unique dataset and make it publicly available. At the same time, we propose an extension of the ISO/TS 15066~\cite{ISOTS15066} equations and demonstrate better predictions of collision forces when protective covers are used.

\subsection*{Contributions}
Therefore, based on the presented state of the art, the contributions of this paper are:
\begin{itemize}
    \item We performed in total 2250 collision measurements on two collaborative robots and on one traditional industrial robot, studying impact force, contact duration, clamping force, and impulse. The dataset, which can be used to develop alternative models of human-robot collisions, is publicly available at \url{https://osf.io/gwdbm}.
    \item We isolated the effects of the passive padding and the active contribution of the sensor to the robot reaction.
    \item We systematically studied the effects of additional parameters: safety stop settings, robot collision reaction, impact direction, and end effector velocity. We present insights into the interplay of these parameters and emphasize that empirical \textit{in situ} measurements are indispensable.
    \item We relate the empirical measurements in different settings to the simplified prescriptions by ISO/TS~15066~\cite{ISOTS15066} and present an extension that takes the stiffness of the protective skin and its compressible thickness into account, leading to more accurate predictions of impact forces.
    
    \item We isolate the potential of individual settings to boost productivity of a collaborative application. 
    
\end{itemize}

\section{Materials and Methods}
\label{sec:materials_methods}
In the following sections, we introduce the necessary concepts for the paper: the robots, the safety stop categories, the characteristics of collisions, the measuring device, the setup used to simulate transient contact, the specifics of our collision evaluation approach, the AIRSKIN safety cover, and the experimental setup itself.

\subsection{Robots}
\label{subsec:robots}
\begin{figure*}[ht]
\begin{subfigure}{0.33\textwidth}
\includegraphics[width=\textwidth]{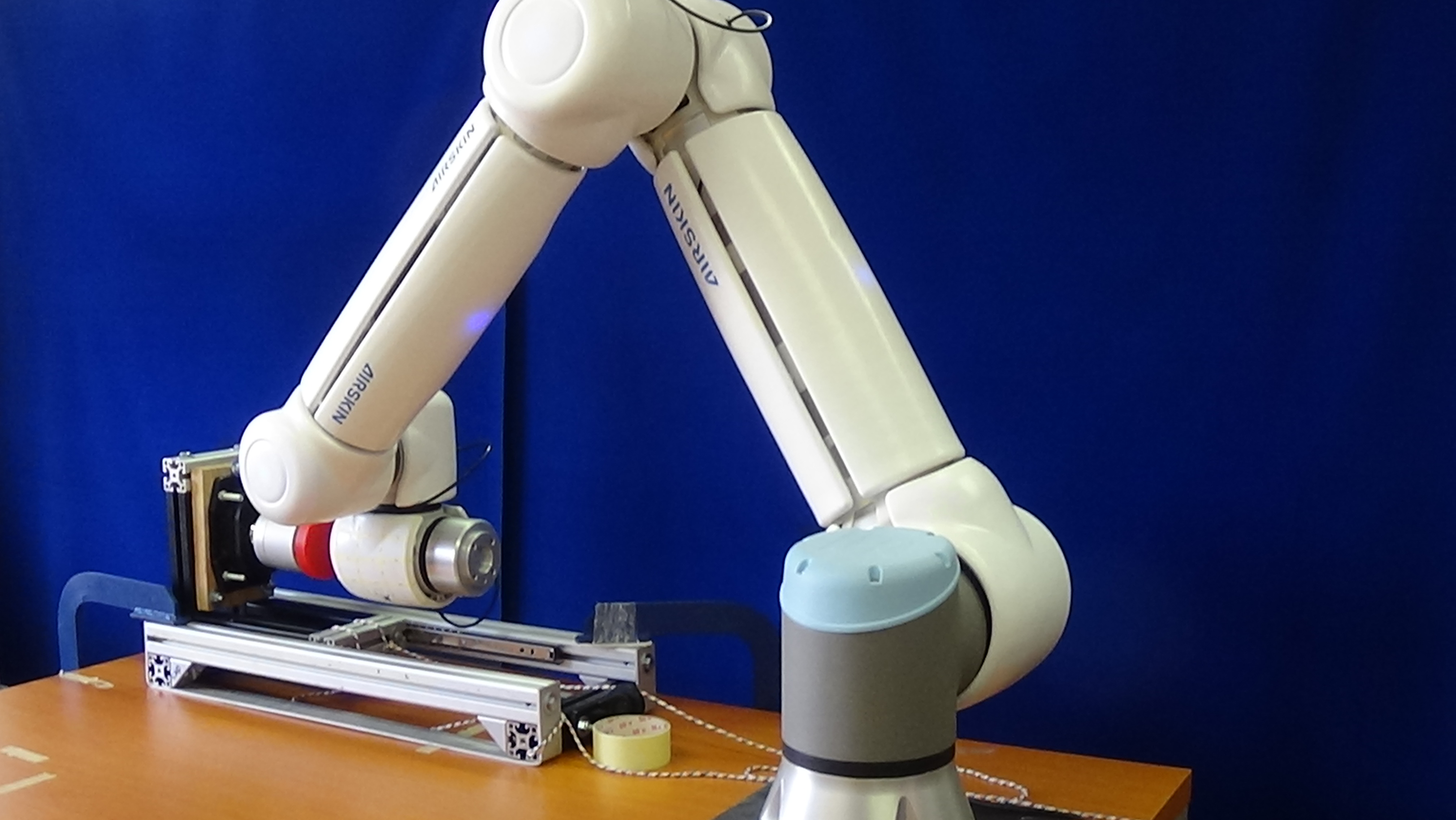} 
\begin{minipage}[t][0.3cm][t]{\linewidth}
\vspace*{-0.2cm}
\caption{UR10e with AIRSKIN\\ (transient contact).}
\label{fig:ur_setup}
\end{minipage}
\end{subfigure}
\hfill
\begin{subfigure}{0.33\textwidth}
\includegraphics[width=\textwidth]{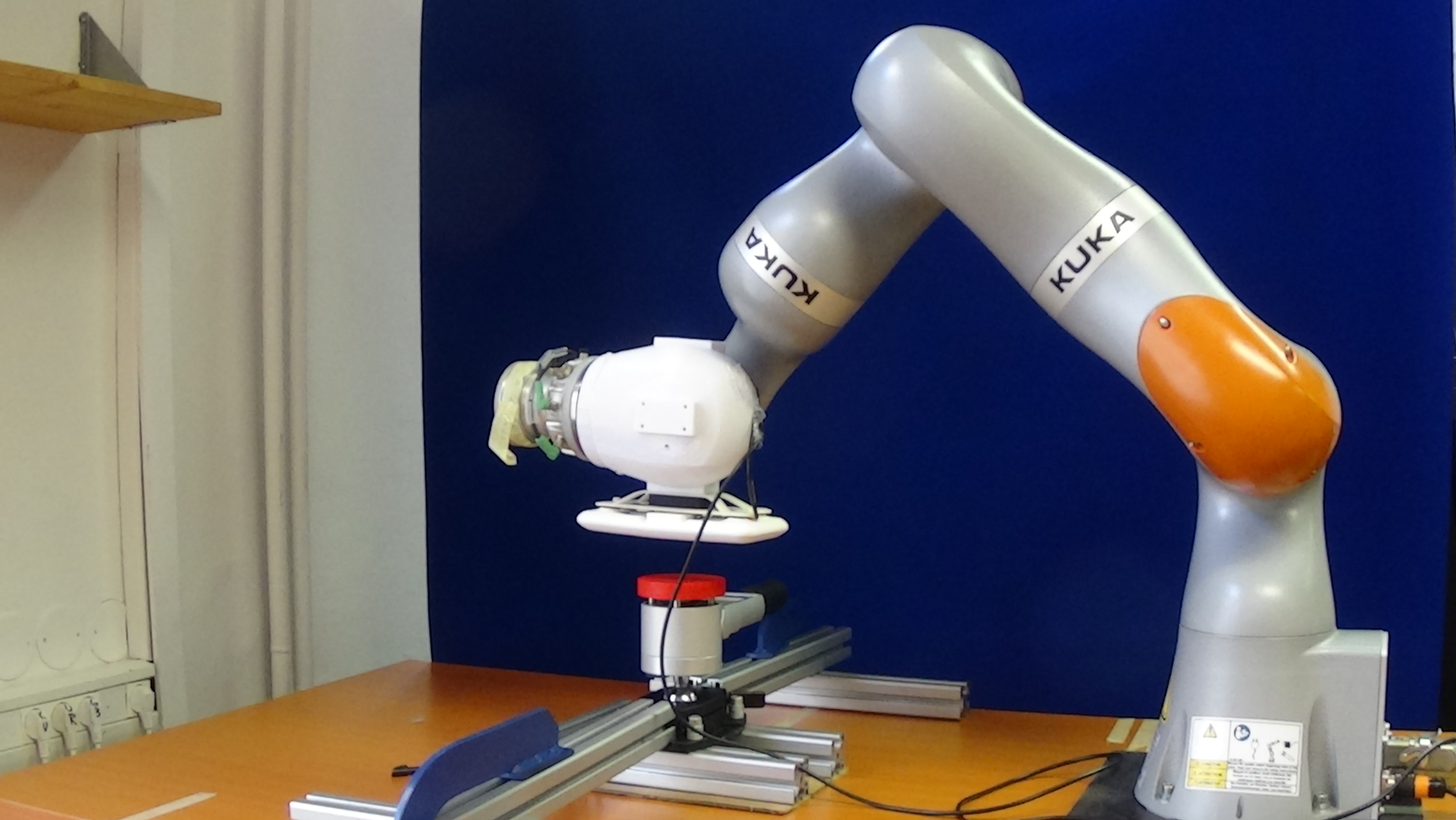}
\begin{minipage}[t][0.3cm][t]{\linewidth}
\vspace*{-0.2cm}
\caption{KUKA iiwa with AIRSKIN module pad\\ (quasi-static contact).}
\label{fig:kuka_setup}
\end{minipage}
\end{subfigure}
\hfill
\begin{subfigure}{0.33\textwidth}
\includegraphics[width=\textwidth]{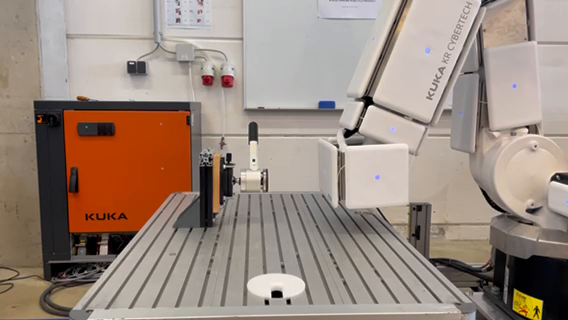}
\begin{minipage}[t][0.3cm][t]{\linewidth}
\vspace*{-0.2cm}
\caption{KUKA Cybertech with AIRSKIN module pad\\ (quasi-static contact).}
\label{fig:ct_setup}
\end{minipage}
\end{subfigure}
\caption{Experiment setup --- robots and the impact measuring device in both impact scenarios.} \label{fig:setup}
\end{figure*}

We used two collaborative and one classical industrial robot for the experiments and controlled them through their standard interfaces. Each robot also has specific safety settings (see below and in Tables \ref{tab:stops} and \ref{tab:safety}).

All robots were using the Cartesian linear movement---where the end effector follows a straight line---toward the impact. Due to various technical limitations (e.g., sensitive equipment on the KUKA iiwa's flange), we measured collisions with the last joint's surface.

\paragraph{Universal robots UR10e (UR10e)} The robot has 6 degrees of freedom (DOF), weighs 33.5~kg, can carry a payload of up to 12.5~kg and has a reach of 1300~mm. The Modbus interface collected the speed and the joint states, safety modes were collected from ROS nodes. Our UR10e is equipped with the protective skin AIRSKIN that adds extra weight (1.8~kg) to the robot, see Fig.\@~\ref{fig:ur_setup}. The skin can be connected to two different safety inputs --- Emergency Stop or Safeguard Stop, see Tab.\@~\ref{tab:stops}. 
The worst-case collection frequency for the robot speed was 800~Hz and 500~Hz for other variables. 

The UR10e robots have four safety presets. We collected data with both the least restrictive (Pre-4) and the second most restrictive safety (Pre-2) presets.\footnote{Least restrictive preset (Pre-4): Allowed power: 1000~W, Momentum: 100~kg~m/s, Stopping time: 1~s, Stopping distance: 2~m, Tool speed: 5.0~m/s, Tool force: 250~N, Elbow speed: 5.0~m/s, Elbow force: 250~N.\newline Second most restrictive (Pre-2): Allowed power: 200~W, Momentum: 10~kg~m/s, Stopping time: 300~ms, Stopping distance: 0.3~m, Tool speed: 0.75~m/s, Tool force: 120~N, Elbow speed: 0.75~m/s, Elbow force: 120~N.}

\paragraph{KUKA LBR iiwa 7 R800 (KUKA iiwa)} This robot has 7 DOF, with the weight of 22.3~kg, a payload of up to 7~kg and a reach of 800~mm, see Fig.\@~\ref{fig:kuka_setup}. Two Java applications controlled the KUKA iiwa robot and collected the relevant data from the robot (1000~Hz frequency). 

The KUKA iiwa assures its safety by monitoring the maximum allowed external torque with joint torque sensors in each joint. We used one external torque limit setting, namely 10~Nm. It was either turned on or turned off. Three different safety stops (Stop 0, Stop 1, Stop 1 op) can be triggered either by the external torque monitor or the AIRSKIN pad.

\paragraph{KUKA Cybertech KR 20 R1810-2 (Cybertech)}
The last robot is a classical industrial robot with 6 DOF, weight of approximately 255~kg, a rated payload of 20~kg, and maximum reach of 1813~mm, see Fig.\@~\ref{fig:ct_setup}. The robot was equipped with AIRSKIN module pads and controlled by a KUKA robot language program. The AIRSKIN module pad can trigger two different safety stops (Stop 1 op, Stop 2). Without sensing capabilities provided by the AIRSKIN, this robot could not be used in a collaborative operation, because the robot would not stop in case of a collision before causing harm to the human collaborator.

\subsection{Safety stops categories}
\label{subsec:stop_cats}
The term ``Stop Category'' refers to the classification of how robot motion is stopped in a safe way based on the standard IEC~60204-1 \cite{IEC60204} and ISO~13850 \cite{ISO13850}. There are four different types (stop category descriptions taken from \cite{URweb}):

\paragraph{Stop Category 0} Robot motion is stopped by immediate removal of power to the robot. It is an uncontrolled stop, where the robot can deviate from the programmed path as each joint brakes as fast as possible. This protective stop is used if a safety-related limit is exceeded or in case of a fault in the safety-related parts of the control system. 

\paragraph{Stop Category 1} Robot motion is stopped with power available to the robot to achieve the stop. Power is removed as soon as the robot stands still.

\paragraph{Stop Category 1 (path maintaining)} Same stop category as the Stop Category 1, but the robot controller has to maintain the pre-planned task path during the controlled stop.

\paragraph{Stop Category 2} A controlled stop with power left available to the robot. The safety-related control system monitors that the robot stays at the stop position.

However, as illustrated in Table~\ref{tab:stops}, the actual stop types of the individual robots do not exactly match those prescribed by the standard and only some of them can be triggered externally like from the protective skin. For the UR10e, Stop 0 can only be triggered through Limit violation and Fault detection and was thus not used in our experiments The Cybertech industrial robot had only Stop 1 op and Stop 2 available. In what follows, we refer to the Emergency stop for UR10e, Stop 0 for KUKA iiwa and Stop 1 op for the Cybertech as the strictest stopping behaviors.

\begin{table}[htb]
\centering
\begin{tabular}{c|ccc}
\textbf{\begin{tabular}[c]{@{}c@{}}Stop\\ Category\end{tabular}} & \textbf{UR10e} & \textbf{\begin{tabular}[c]{@{}c@{}}KUKA\\ iiwa\end{tabular}} & \textbf{\begin{tabular}[c]{@{}c@{}}KUKA\\ Cybertech\end{tabular}} \\ \hline
\multirow{2}{*}{Stop 0} & \begin{tabular}[c]{@{}c@{}}Limit \\ violation\end{tabular} &\cellcolor{gray!30} \multirow{2}{*}{Stop 0} & \multirow{2}{*}{Stop 0} \\
 & \begin{tabular}[c]{@{}c@{}}Fault \\ detection\end{tabular} &\cellcolor{gray!30}  &  \\ \hline
Stop 1 & ----- &\cellcolor{gray!30} Stop 1 & Stop 1 \\ \hline
\begin{tabular}[c]{@{}c@{}}Stop 1 (path\\ maintaining)\end{tabular} &\cellcolor{gray!30} \begin{tabular}[c]{@{}c@{}}Emergency \\ stop (E-stop)\end{tabular} &\cellcolor{gray!30} \begin{tabular}[c]{@{}c@{}}Stop 1 op\end{tabular} &\cellcolor{gray!30} \begin{tabular}[c]{@{}c@{}}Stop 1 op\end{tabular} \\ \hline
Stop 2 & \cellcolor{gray!30} \begin{tabular}[c]{@{}c@{}}Safeguard \\ stop (S-stop)\end{tabular} & ----- & \cellcolor{gray!30} Stop 2
\end{tabular}
\caption{Stop categories comparison between robots. Gray values indicate stops categories which can be triggered externally (i.e., AIRSKIN can trigger it). \label{tab:stops}}
\end{table}

\begin{table}[htb]
\centering
\begin{tabular}{c|c|ccc}
\textbf{Robot} & \textbf{Safety} & \multicolumn{3}{c}{\textbf{Values}} \\ \hline
\multirow{2}{*}{UR10e} & Preset & Pre-2 & Pre-4 &  \\  \cline{2-5}
 & Skin & E-Stop & S-Stop &  \\ \hline
\multirow{2}{*}{\begin{tabular}[c]{@{}c@{}}KUKA\\ iiwa\end{tabular}} & \begin{tabular}[c]{@{}c@{}}External\\ torque\end{tabular} & Stop 0 & Stop 1 & Stop 1 op \\ \cline{2-5}
 & Skin & Stop 0 & Stop 1 & Stop 1 op \\ \hline
\begin{tabular}[c]{@{}c@{}}KUKA\\ Cybertech\end{tabular} & Skin & Stop 1 op & Stop 2 & 
\end{tabular}
\caption{Safety settings overview. The possible combinations of triggered stops and trigger origins or robot specific safety settings.}
\label{tab:safety}
\end{table}

\subsection{Collisions, their modeling, and ISO/TS 15066}
\label{subsec:nature}

\begin{figure}[htb]
	\centering
	\includegraphics[width=0.4\textwidth]{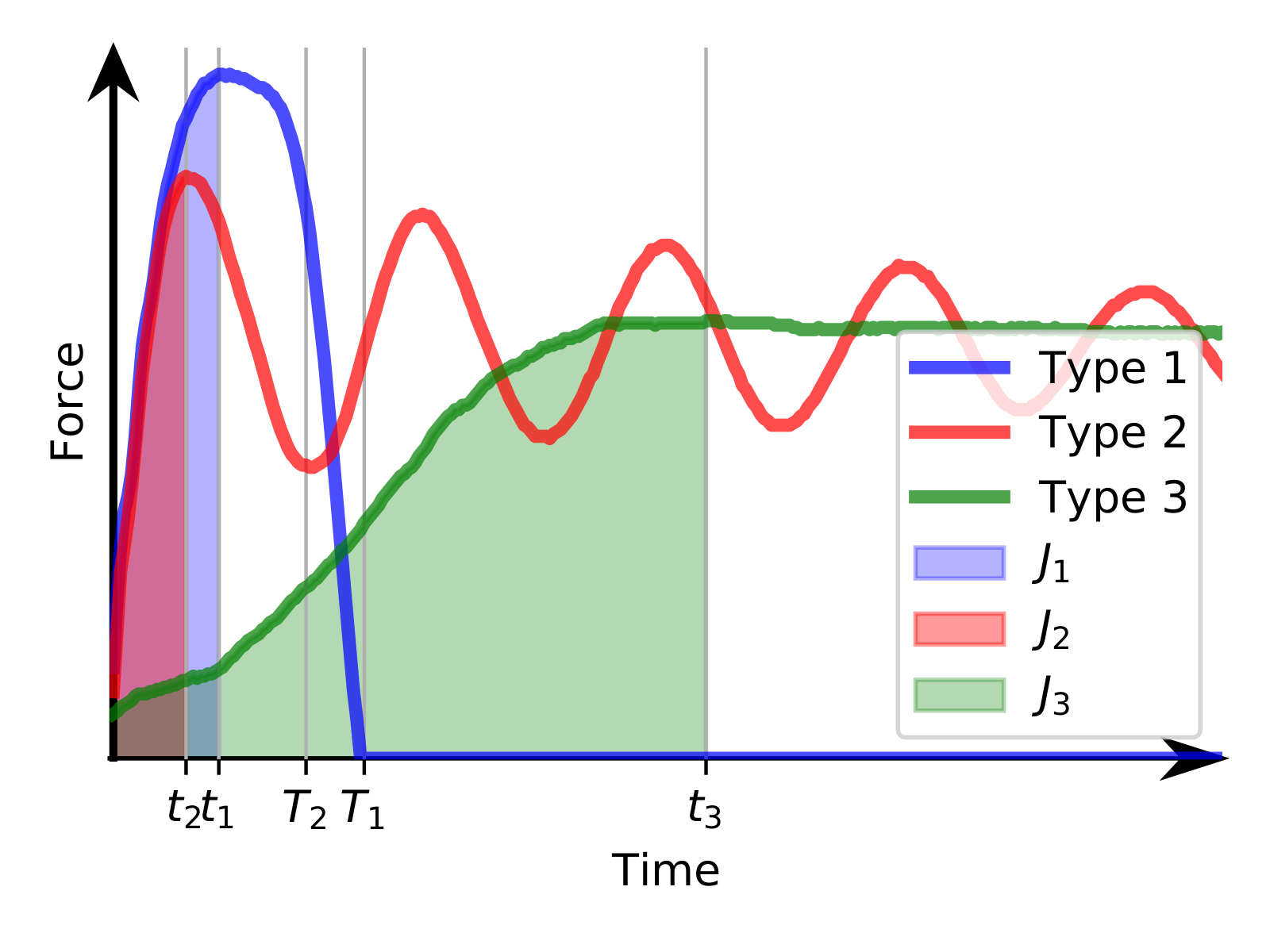}
	\caption{Collision phases and measurement times. We present three types of collisions, unconstrained dynamic impact (Type 1) and constrained dynamic impacts with clamping (Type 3), and oscillation (Type 2). The time instants on the x-axis have the following meaning: $t_{1}$, $t_{2}$, $t_{3}$ mark the point of peak impact force while $T_{1}$, $T_{2}$ mark the respective dynamic collision durations. The areas $J_{1}$, $J_{2}$, $J_{3}$ denote the respective impulses for each type of collision. } 
	\label{fig:phases}
\end{figure}

In order to precisely distinguish between various scenarios of human-robot collisions, we need to analyze the nature of collision events. This was presented in \cite{Vicentini2020a, Haddadin2015}. According to \cite{Haddadin2015}, we can decompose the collision into two phases. Phase I is the initial dynamic impact followed by the Phase II force profile that depends on the clamping nature of the incident. 

Vicentini~\cite{Vicentini2020a} distinguishes three possible impact scenarios:
\begin{itemize}
    \item unconstrained dynamic impact (no force in Phase II),
    \item constrained dynamic impact without clamping (diminishing force in Phase II),
    \item constrained dynamic impact with clamping (force is not diminishing in Phase II).
\end{itemize}

However, based on our empirical study, we present three slightly different scenarios in \fig{\ref{fig:phases}}. While Type 1 represents an unconstrained dynamic impact and Type 3 matches the constrained dynamic impact with clamping, Type 2 marks a different case than was presented in the literature where oscillation occurs in Phase II. From an overall perspective, the trend of the oscillation could equate it with one of the two constrained scenarios. Nonetheless, the amplitude can make the oscillation unacceptable from the perspective of the safe force limits.

The TS~15066 presents only two scenarios, a \textit{transient} contact, i.e., a dynamic impact that is unconstrained or it is constrained without clamping, and \textit{quasi-static} contact, i.e., dynamic impact with clamping. 

An integral part of determining the safety of the interaction is finding the appropriate safe velocities and impact forces. This is covered by the equation A.6 from TS~15066 relating velocity and impact force for transient contact:
\begin{equation}
    v \leq  \frac{F_\mathrm{max}}{\sqrt{k}}\sqrt{m_{R}^{-1} + m_{H}^{-1}} = \frac{F_\mathrm{max}}{\sqrt{k\cdot \mu}},
    \label{eq:v_pfl_orig}
\end{equation}

where $m_R$ is the effective robot mass, $m_H$ is the human body part mass, $\mu = (m_{R}^{-1} + m_{H}^{-1})^{-1}$ is the reduced mass of the two-body system, $k$ is the spring constant for the human body part, and $F_\mathrm{max}$ is the maximum impact force permitted for the given body region---280~N (first 0.5~s of the impact) or 140~N (after 0.5~s of the impact) for the back of the non-dominant hand. The mass of the robot $m_R$ is given by the used robot, while $m_H$ is given by the scenario. For transient contact, $m_H$ is 5.3~kg---the mass of the measuring device.

If we investigate constrained dynamic impacts, even without clamping, we can approximate $m_H^{-1}\approx 0$ as in \cite{Lucci2020}\footnote{The impacted body part is constrained and thus immovable. Its weight in the PFL two-body spring model can therefore be considered as significantly larger than the other body's, and hence approximated as infinite.}. This approximation allows us also to simplify the situation by investigating the relative velocity as the robot velocity with the human hand being still: 
\begin{equation}
    v \leq  \frac{F_\mathrm{max}}{\sqrt{k\cdot m_R}}.
    \label{eq:v_pfl}
\end{equation}

TS~15066 allows considering the ``effective robot mass'' statically as $M/2 + m_L$, a function of the total mass of the moving parts of the robot $M$, and the effective payload $m_L$. In the case of the collaborative robots used in this work, the moving masses $M$ of the UR10e and KUKA iiwa, are approximately 30~kg and 20~kg, respectively. These values, together with other variables set based on TS~15066~\cite{ISOTS15066} ($m_L=\ $ 0~kg, and $k=\ $75000~N/m; see Eq.~\ref{eq:v_pfl}) result in a permissible velocity up to 0.26~m/s for the UR10e and 0.32~m/s for KUKA iiwa if the force limit for the first 0.5~s after the collision (280~N) is considered. For the period after the first 0.5~s (clamping scenario), the stricter limit of 140~N applies, resulting in prescribing a maximum velocity of 0.13~m/s (UR10e) / 0.16~m/s (KUKA iiwa).

Khatib~\cite{khatib1995inertial} introduced the robot effective mass as a dynamic property depending on the robot's configuration and the impact direction. This has been later adopted by many others, sometimes also called reflected mass (e.g., \cite{Haddadin2016,Mansfeld2018,Lucci2020,Haddadin2015,Lee2013}).
The effective mass of a manipulator in a given direction $\boldsymbol{u}$ can be modeled using the formula \cite{Lee2013}:
\begin{equation}
    m_{\boldsymbol{u}}^{-1} = \boldsymbol{u}^T [J(\boldsymbol{q}) M^{-1}(\boldsymbol{q}) J^{T}(\boldsymbol{q})] \boldsymbol{u},
    \label{eq:eff_mass}
\end{equation}
where $\boldsymbol{q}$ are the joint angles of a given position, $M(\boldsymbol{q})$ and $J(\boldsymbol{q})$ are the inertia matrix and the Jacobian matrix of the manipulator, respectively (see, e.g., \cite[Ch. 3 and Ch. 7]{Siciliano2010}).

\subsection{Modeling collisions with soft protective cover}
\label{subsec:model_soft}
Eq.~\ref{eq:v_pfl_orig} is derived from the assumption that the entire kinetic energy of the robot is transferred into spring energy of the respective human body region in a fully inelastic contact (see Eq.~A.2 from TS~15066):
\begin{equation}
    E = \frac{\mu\cdot v^2}{2} = \frac{F^2}{2\cdot k}
    \label{eq:pfl_energies}
\end{equation}
This is, however, not the case here, where the (passive or active) soft protective cover also stores spring energy. This spring energy is given by the spring constant (stiffness) of the soft protective material and the amount it can compress before being completely flattened out.
\begin{equation}
    E = \frac{\mu\cdot v^2}{2} = \frac{F^2}{2\cdot k} + \frac{d_s^2\cdot k_s}{2},
    \label{eq:pfl_energies_soft}
\end{equation}
where $d_s$ is the compressible thickness of the soft protective material, and $k_s$ its spring constant.
This increases the permissible speed:
\begin{equation}
    v \leq  \sqrt{\frac{F_\mathrm{max}^2}{k\cdot \mu} + \frac{d_s^2\cdot k_s}{\mu}}
    \label{eq:v_pfl_soft}
\end{equation}
The stiffness of a pad can vary over its surface, e.g., an AIRSKIN module pad is harder right over the electronics housing (see also Fig.~\ref{fig:kuka_stif}). Let us approximate the stiffness at this location with a spring constant of $k_s =\ $3000~N/m and a compressible thickness of $d_S =\ $16~mm. This represents the maximum compressible thickness for the AIRSKIN modules, which we are using for simplicity instead of the actual compression at the location.

For the collaborative robots equipped with AIRSKIN (see Sec.~\ref{subsec:airskin} for details) and using Eq.~\ref{eq:v_pfl_soft}, this gives the following permissible velocities for quasi-static contact. During the first 0.5~s after impact (280~N force limit), the permissible velocity for the UR10e goes up from 0.26~m/s to 0.35~m/s; for KUKA iiwa, from 0.32 to 0.43~m/s. For the period after the first 0.5~s (clamping scenario, 140~N limit), the maximum permissible velocity goes up from 0.13 to 0.26 and from 0.16 to 0.32~m/s for the UR10e and KUKA iiwa, respectively---thanks to the energy absorption by the soft protective material.

Eq.~\ref{eq:v_pfl_soft} can be rearranged to express force as follows:
\begin{equation}
    F =  \sqrt{\left(v^2 - \frac{d_s^2\cdot k_s}{\mu}\right)\cdot (k\cdot \mu)}
    \label{eq:f_pfl_soft}
\end{equation}
For small velocities, it may be that $v^2 < (d_s^2\cdot k_s)/\mu$. In these cases, no prediction is available (Fig.~\ref{fig:ur_summary} and~\ref{fig:kuka_summary}). Here, actual rather than maximum skin compression should be considered---this is not used in this work.
If the stiffness and compression of the protective skin is not considered, the term $(d_s^2\cdot k_s)/\mu$ is 0 and the equation is equivalent to the corresponding TS 15066 formulas (Sec.~\ref{subsec:nature}).

This gives four variants of Eq.~\ref{eq:f_pfl_soft} which we will use later (Fig.~\ref{fig:ur_summary} and \ref{fig:kuka_summary}): (i) TS 15066 and quasi-static contact, (ii) TS 15066 and transient contact, (iii) modified TS 15066 and quasi-static contact, and (iv) modified TS 15066 and transient contact. For quasi-static contact (i) and (iii), the simplification $m_H^{-1}=0$ cited above is used. For (i) and (ii), the skin stiffness is not modeled; hence $(d_s^2\cdot k_s)/\mu = 0$.

\subsection{Measuring device}
\label{subsec:measurement_device}
We used the \textit{CBSF-75-Basic} impact measuring device designed for validation of collaborative applications of robots. It allows the measuring of forces in the range from 20~N up to 500~N with a certified measurement error up to 3~N. The measurement collection frequency is 1000~Hz and it starts as soon as a 20~N impact force threshold is reached and thereafter continues for 5 seconds. Therefore the pre-threshold force evolution is not collected.

In accordance with TS~15066, appropriate K1 damping materials are added to the device in order to simulate the properties of the human body region. Namely, we mimic impacts on the back of the non-dominant hand by using the Basic 75~N/m device and the damping material with the hardness Sh 70.

\subsection{Transient contact simulation apparatus}
\begin{figure}[ht]
\begin{subfigure}{0.23\textwidth}
\includegraphics[width=\textwidth]{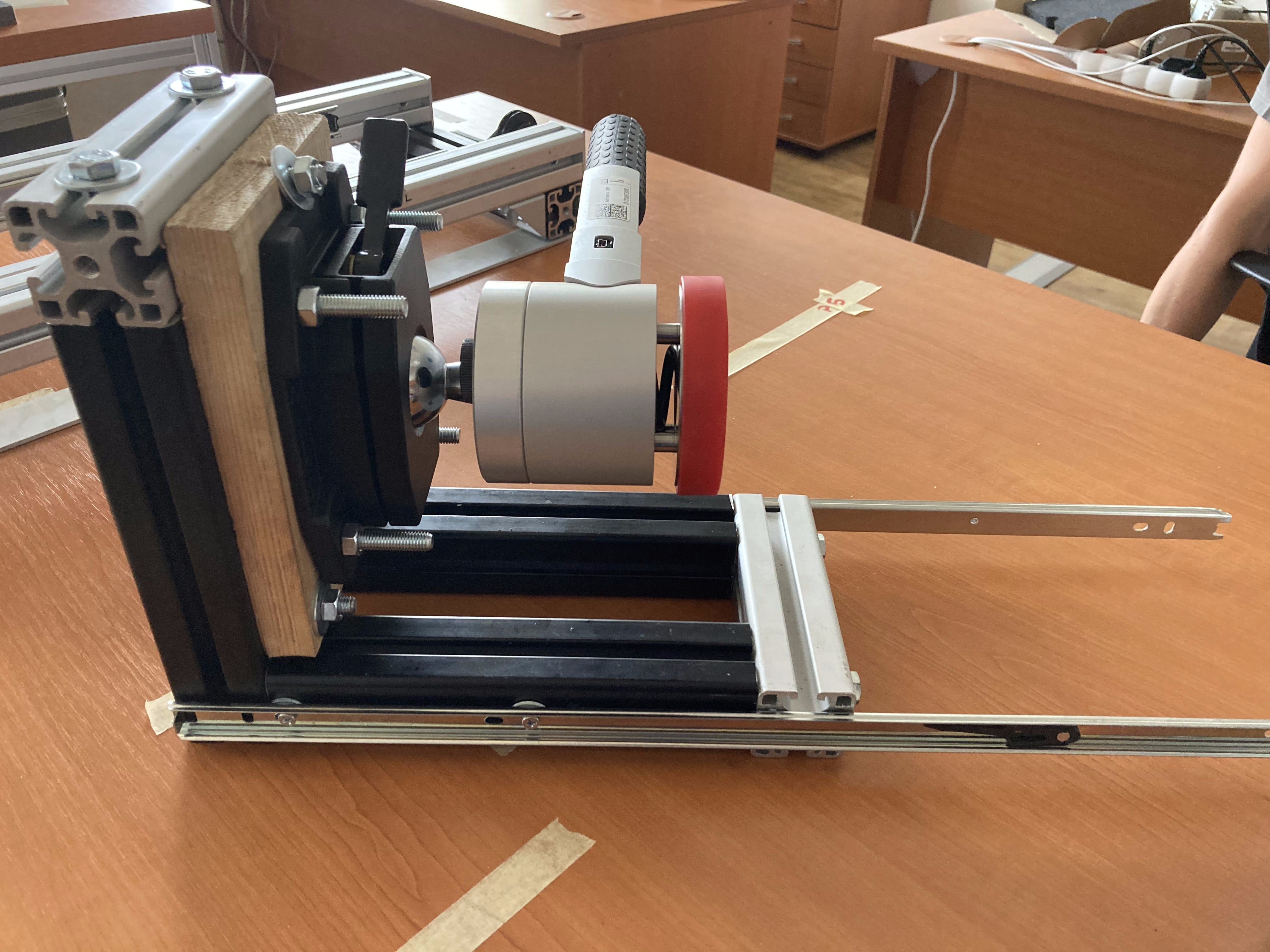} 
\caption{Moving mass}
\label{fig:tr_mass}
\end{subfigure}
\hfill
\begin{subfigure}{0.23\textwidth}
\includegraphics[width=\textwidth]{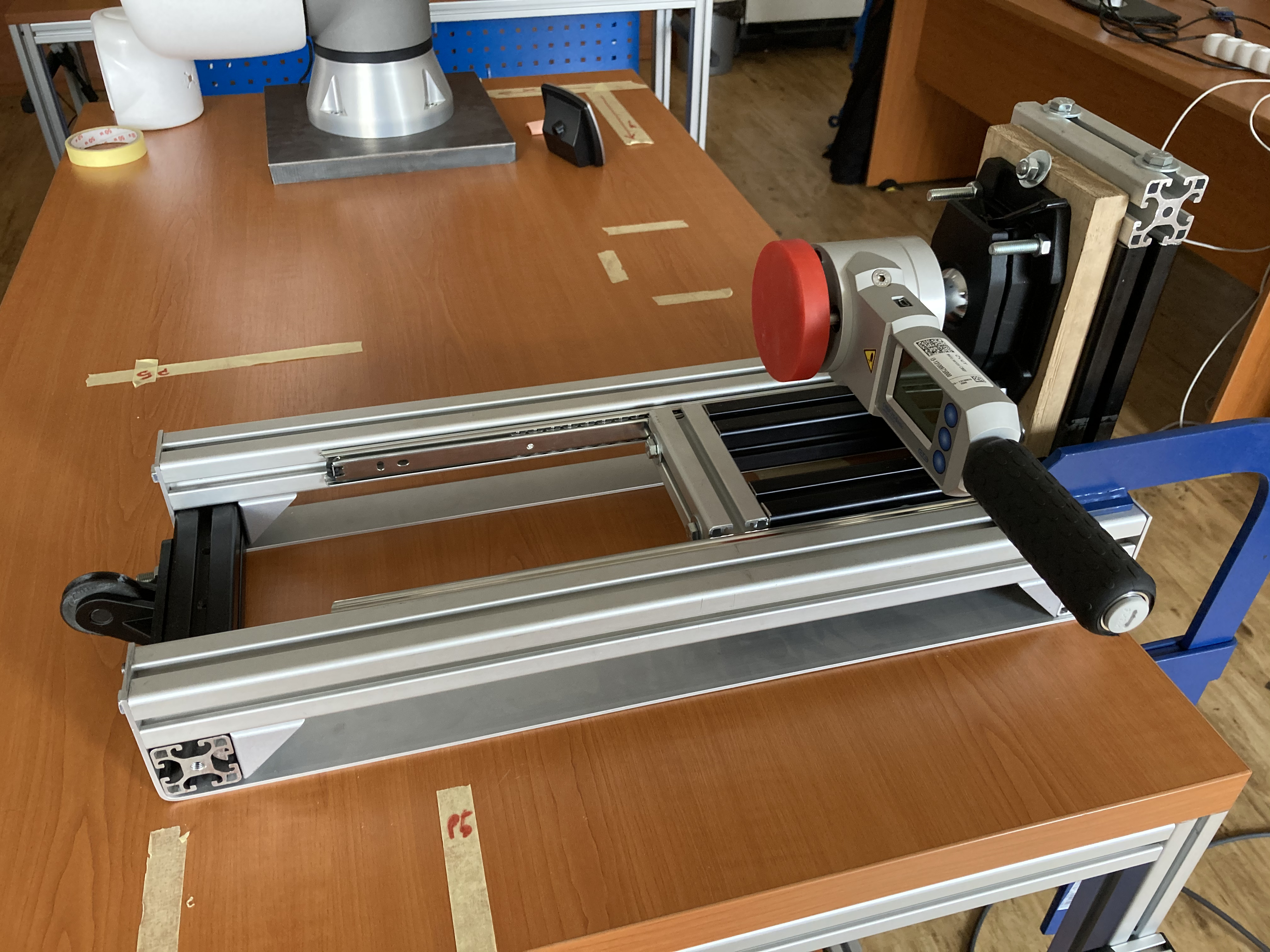}
\caption{Whole construction}
\label{fig:tr_const}
\end{subfigure}
\caption{Transient contact simulation construction.}
\label{fig:transient_device}
\end{figure}

We designed and assembled a custom mechanism to simulate \textit{transient} contact with a frame made of aluminum profiles. The structure can be divided into two parts: a moving mass with the measuring device (see Fig.\@~\ref{fig:tr_mass}), and a static base with ball bearing drawer slides (whole construction in Fig.~\ref{fig:tr_const}). 

To secure similar conditions between individual transient experiments, we assessed the key characteristics of the apparatus. We decided to experimentally determine the smallest applied force ($F_f$) enough to move the moving mass horizontally and keep it the same for all experiments. Moreover, since this force affects the measured force by the measuring device, we wanted to minimize it as much as possible. Then we can assume the measured force represents a friction force and we can compute the relevant friction coefficient. For the sake of simplicity, we assume that the ball bearing drawer slides cause static friction instead of the more precise rolling resistance. The coefficient of static friction $\mu$ can be computed as $F_f = \mu_0 \cdot F_N$, where $F_f$ is the static friction force, and $F_N$ is the normal force. The moving mass weight (together with the measuring device) was 5.3~kg. Thus the normal force between the moving mass and the static base is approximately $F_N$ = 52~N, the measured friction force is $F_f$ = 3.2~N, and the coefficient of static friction is $\mu_0 = 0.062$.

\subsection{Collision evaluation}
To analyze the skin effect on the safety of the interaction, several physical quantities were investigated. Naturally, the effect on impact forces should be examined because, as mentioned before, PFL mode deals with collisions by prescribing the maximum exerted forces. Forces were compared in both phases---peak impact force in Phase I (initial dynamic impact) and clamping force in Phase II.
The skin effect on initial impact duration should also be examined, because we assume the cover prolongs the collision time. 
This would be the same principle as with airbags which distribute the collision impulse into a longer period of time and thus diminish the maximum force applied on the colliding human body part (head in the case of airbags)~\cite[Ch. 6.15]{Hamm2020}.

Based on these two physical quantities, the impulse can be calculated as the integral of the measured impact force. For us, the impulse from collision start to peak impact force is practical for further use. Based on the impulse-momentum theorem (the change in momentum equals the impulse applied to it), the impulse $J$ (in $N \cdot s$) can also be computed as:
\begin{equation}
    J = \int_{0}^{t_p}F \mathop{}\!d t = \Delta p =  m_r \Delta v = m_r v_0,
    \label{eq:imp}
\end{equation}
where $t_p$ is the time of peak impact force, $F$ is measured impact force, $\Delta p$ is the change in momentum, $m_r$ is the robot mass, $\Delta v$ is difference between the initial velocity $v_0$ before collision and the velocity at time $t_p$ which, we assume, is equal to zero. The impulses, peak impact forces, and relevant peak impact times of various force profiles are also shown in \fig{\ref{fig:phases}}. As can be seen, the relation can be used to compute the expected robot mass $m_r$: 

\begin{equation}
    m_r = \frac{J}{v_0},
    \label{eq:mass}
\end{equation}
where J is the computed impulse from collision start to peak impact force and $v_0$ is the initial velocity before collision.

For evaluation of the collision, we tracked also the reaction times, which is possible only with the UR controller. We measured the time between the onset of the collision, i.e., the measuring device starts recording as the force exceeds its 20~N threshold, and the UR controller issuing a stop status. The KUKA iiwa suspends logging when a safety response is triggered.

\subsection{AIRSKIN safety cover and collision sensor}
\label{subsec:airskin}
All of the robots are equipped with AIRSKIN, a soft pressure-sensitive collision sensor that covers the robot. AIRSKIN is a commercial product available for a number of robots, in a price range of 9000--16000 EUR depending on the robot model. It has a guaranteed lifetime of 10 years. AIRSKIN is made of individual pads, where each pad consists of an airtight hull covering a soft support structure and pressure sensors placed inside the hull~\cite{Zillich2012}. Deformation of the pad as a result of contact leads to an increase in internal pressure. An increase beyond a configurable threshold (100~Pa) issues a stop signal from the sensor. Gradual pressure changes due to atmospheric changes or temperature changes of the pad are filtered out. Furthermore, each pad is slightly pressurized to 600~Pa over atmospheric pressure. Damage of the hull leads to a drop of the internal pressure, which causes the pad to go into an error state, also issueing a stop signal from the sensor.

All the electronics (2 ARM-based microcontrollers, 2 sets of internal and external MEMS pressure sensors, piezo-electric pump, and valve) are contained on the PCB mounted in each pad. All pads are connected in series, acting as opening switches for 2 safety channels, that are connected to the safety inputs of the robot controller.

\begin{figure}[ht]
    \begin{subfigure}{0.23\textwidth}
    	\includegraphics[width=\textwidth]{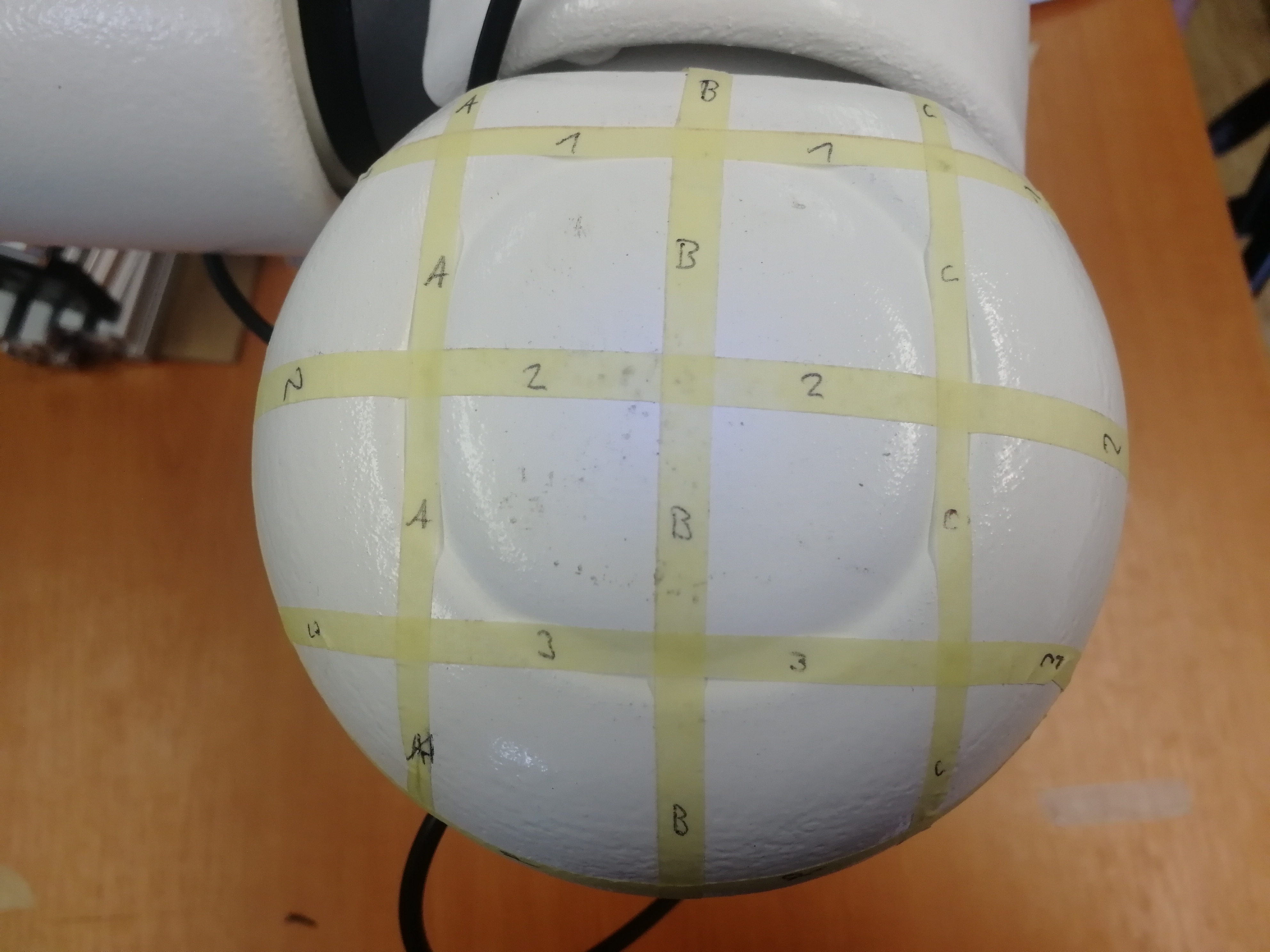}
    	\caption{UR-skin}
    	\label{fig:ur_pad}
    \end{subfigure}
\hfill
\begin{subfigure}{0.23\textwidth}
	\includegraphics[angle=90,width=\textwidth]{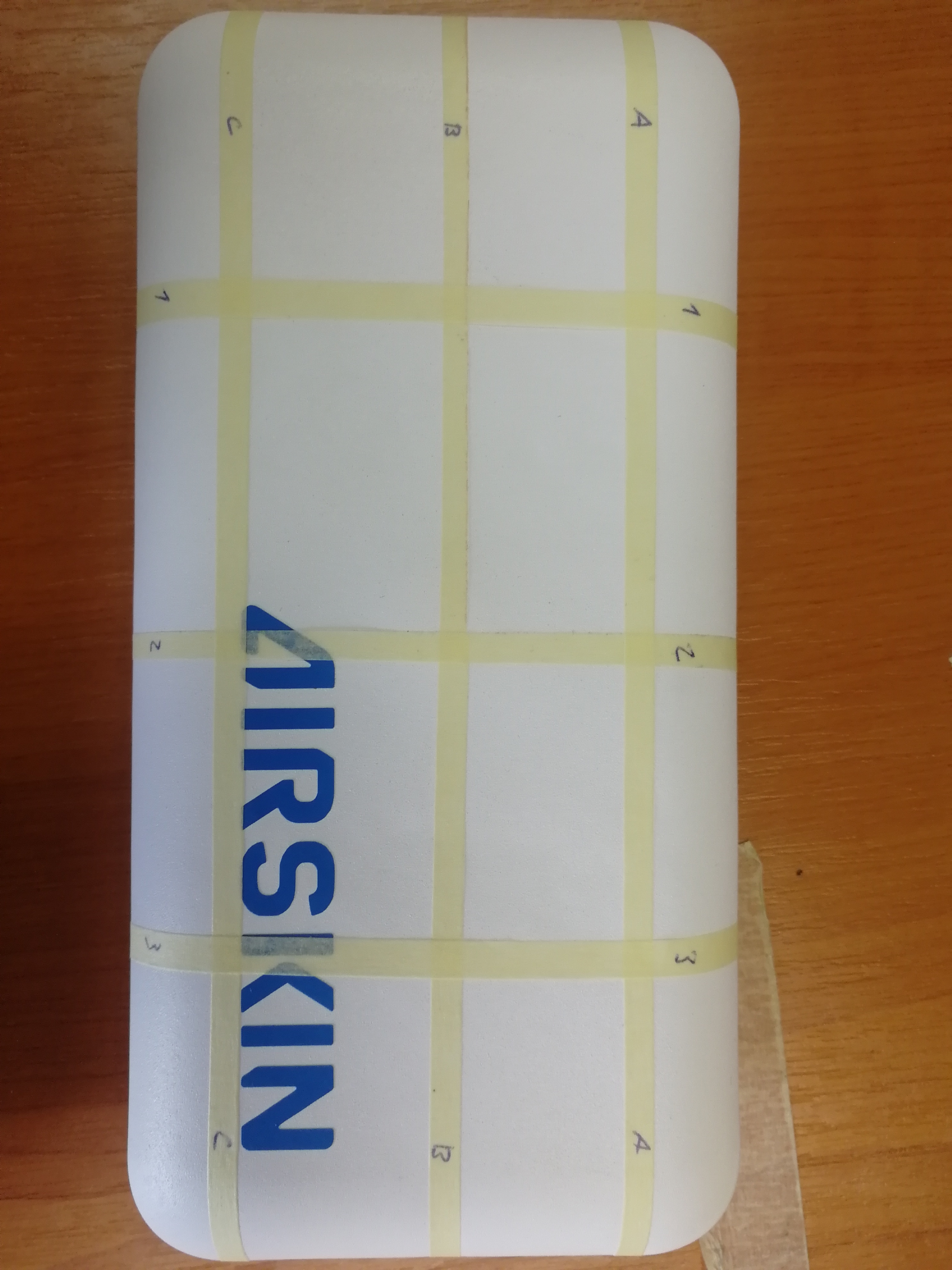}
	\caption{The Pad}
	\label{fig:kuka_pad}
\end{subfigure}
\caption{AIRSKIN pads with measurement grid.}
\label{fig:pads}
\end{figure}

These pads can either be in the form of custom shaped pads for the given robot (in our case for the UR10e robot, UR-skin in what follows; see Fig.\@~\ref{fig:ur_pad}) or in the form of rectangular pads, AIRSKIN Module Pads (Pad in what follows; see Fig.\@~\ref{fig:kuka_pad}) used on both KUKA robots in this work. The UR-skin is an earlier model from 2017, while the AIRSKIN module pad is from 2019.

The recommendation from the manufacturer is to use AIRSKIN with Stop Categories 1 and 2 and not with Stop Category 0. Frequent stoppage in Stop Category 0 could damage the robot.

We wanted to verify the properties of the sensors to know if their results are comparable. We designed a measuring device that pushed the pads at various locations to study the pads' stiffness. The pushed locations were identified by a matrix --- see  Fig.\@~\ref{fig:pads}. Due to the selected impact locations (point B2 on our matrix on both), the effect of the geometry should be low as both of the impact surfaces are flat. However, we have to take into account the surrounding area of the collision point also as the collision will happen with a flat area (the impact surface of CBSF-75-Basic) and not a single point. In addition, a large difference in the stiffness of the material could influence the measured impact force. 

We used the measuring device also to determine the threshold force that activates the AIRSKIN pads. For this, the measuring device was connected with the AIRSKIN pad safety output. The device then slowly increased the exerted force upon the pad until it was stopped by the pad. The results of these measurements are presented in Sec.~\ref{subsec:airskin_properties}.

\subsection{Experiment setup and data collection}
\label{subsec:setup}
The experiments covered \textit{quasi-static} and \textit{transient} contacts. Each experiment consisted of a series of impacts at similar locations in the workspace of the given robot and at different velocities and different impact directions --- along world frame axes.

\begin{figure*}[ht]
\begin{subfigure}{0.33\textwidth}
\includegraphics[width=\textwidth]{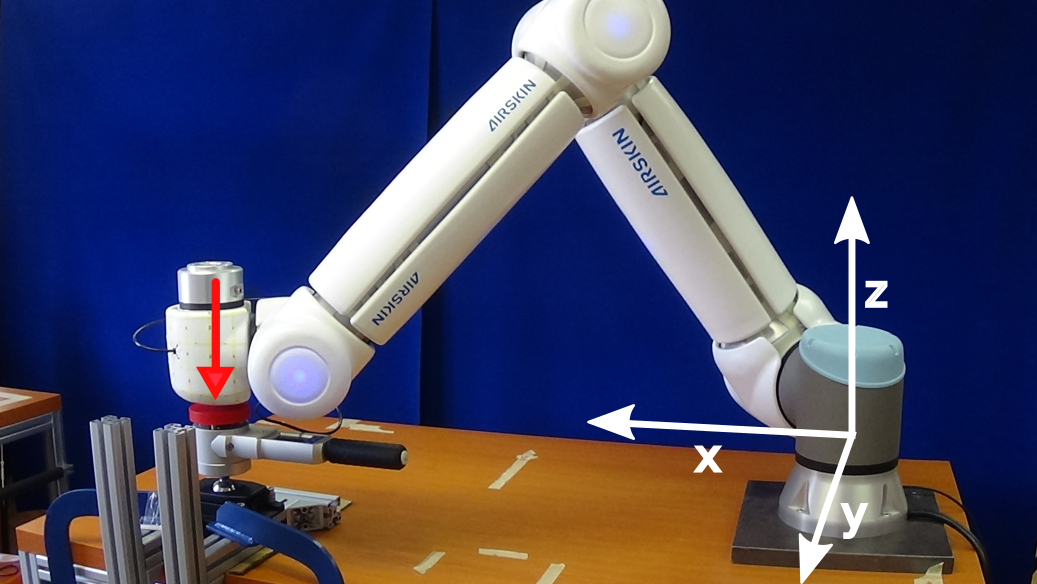} 
\caption{Impact direction downward ($[0,0,-1]$), quasi-static case}
\label{fig:place_downward}
\end{subfigure}
\hfill
\begin{subfigure}{0.33\textwidth}
\includegraphics[width=\textwidth]{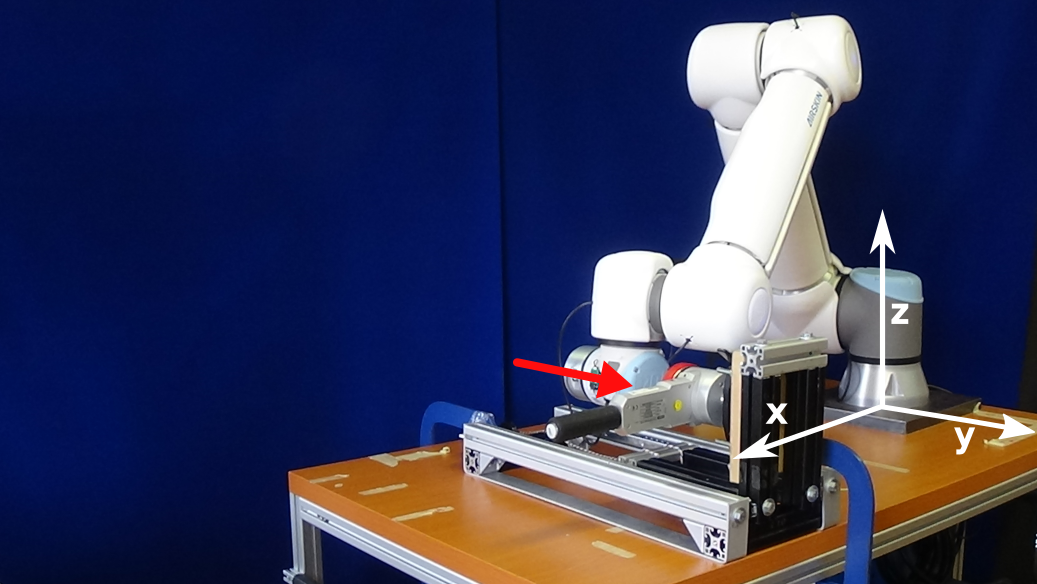}
\caption{Impact direction along the $y$-axis ($[0,1,0]$), transient case}
\label{fig:place_x}
\end{subfigure}
\hfill
\begin{subfigure}{0.33\textwidth}
\includegraphics[width=\textwidth]{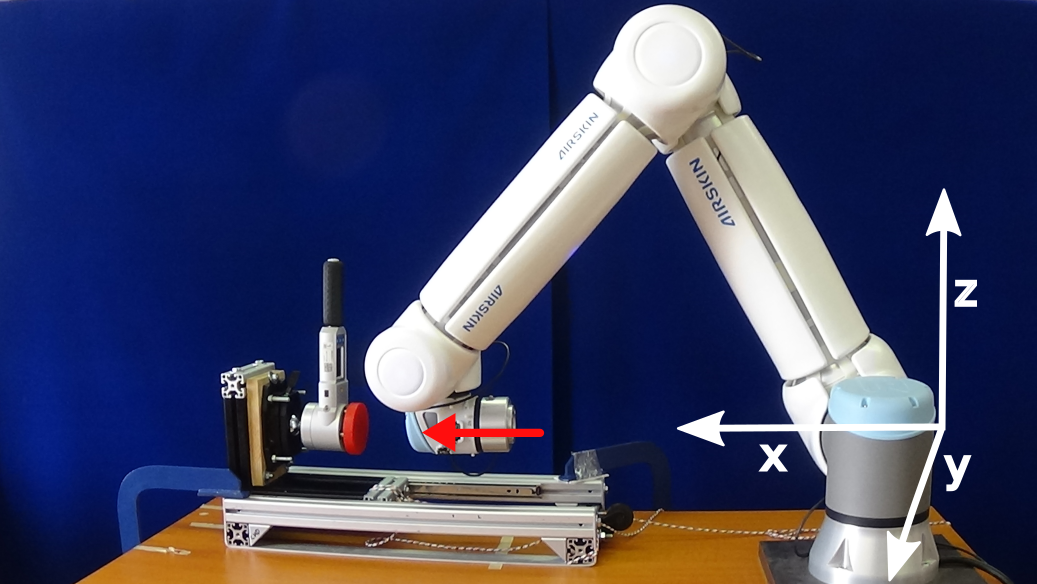}
\caption{Impact direction along the $x$-axis ($[1,0,0]$), transient case}
\label{fig:place_y}
\end{subfigure}
\caption{Impact directions. The origin of the world reference frame is always at the base frame of the robot. The coordinates in the image are moved for visibility.}
\label{fig:places}
\end{figure*}

As the robot world-frames are different, we use a right-handed coordinate frame of reference at the robot base with the $y$-axis along the short side of the table, the $x$-axis along the longer side of the table, and $z$-axis perpendicular to this plane --- see \fig{\ref{fig:places}}. In this coordinate frame, our impact directions are downward ($[0,0,-1]$), along the $x$-axis ($[1,0,0]$), and along the $y$-axis ($[0,1,0]$). 

We defined 5 impact \textit{places}---where the \textit{place} refers to location in the workspace, impact direction, and contact type---for each robot: 3 for quasi-static contacts (downward, along $x$-axis, along $y$-axis) and 2 for transient contacts (along $x$-axis, along $y$-axis). Their world coordinates can be found in Tab.\@~\ref{tab:places}. The locations of every place for a given direction are similar for both types of contacts, making a comparison possible.

\begin{table}[htb]
\centering
\resizebox{250pt}{!}{%
\begin{tabular}{ccc|ccc}
\multicolumn{3}{c|}{\textbf{Place}} & \multicolumn{3}{c}{\textbf{Coordinates {[}m{]}}} \\ \hline
\textbf{Type} & \textbf{\#} & \textbf{\begin{tabular}[c]{@{}c@{}}direction \\ vector\end{tabular}} & \textbf{UR10e} & \textbf{\begin{tabular}[c]{@{}c@{}}KUKA \\ iiwa\end{tabular}} & \textbf{\begin{tabular}[c]{@{}c@{}}KUKA \\ Cybertech\end{tabular}} \\ \hline
\multirowcell{3}{quasi-\\static} & 0 & $\begin{pmatrix}\phantom{-}0\\ \phantom{-}0\\  -1\end{pmatrix}$ & $\begin{matrix}\phantom{-}0.85\\ \phantom{-}0.27 \\  \phantom{-}0.14\end{matrix}$ & $\begin{matrix}\phantom{-}0.66\\ \phantom{-}0.00\\  \phantom{-}0.14\end{matrix}$ & -- \\ \cline{2-6}
 & 1 & $\begin{pmatrix}\phantom{-}0\\ \phantom{-}1\\  \phantom{-}0\end{pmatrix}$ & $\begin{matrix}\phantom{-}0.79\\ \phantom{-}0.14\\  \phantom{-}0.16\end{matrix}$ & $\begin{matrix}\phantom{-}0.35\\ \phantom{-}0.14\\  \phantom{-}0.16\end{matrix}$ & $\begin{matrix}0.00\\ 0.90\\ 0.18\end{matrix}$ \\ \cline{2-6}
 & 2 & $\begin{pmatrix}\phantom{-}1\\ \phantom{-}0\\  \phantom{-}0\end{pmatrix}$ & $\begin{matrix}\phantom{-}0.80\\ -0.22\\  \phantom{-}0.16\end{matrix}$ & $\begin{matrix}\phantom{-}0.37\\ -0.31\\  \phantom{-}0.16\end{matrix}$ & $\begin{matrix}0.25\\0.75\\ 0.18\end{matrix}$ \\ \hline
\multirow{2}{*}{transient} & 3 & $\begin{pmatrix}\phantom{-}0\\ \phantom{-}1\\  \phantom{-}0\end{pmatrix}$ & $\begin{matrix}\phantom{-}0.79\\ \phantom{-}0.18\\  \phantom{-}0.16\end{matrix}$ & $\begin{matrix}\phantom{-}0.35\\ \phantom{-}0.10\\  \phantom{-}0.16\end{matrix}$ & $\begin{matrix}0.00\\ 0.90\\ 0.18\end{matrix}$ \\ \cline{2-6}
 & 4 & $\begin{pmatrix}\phantom{-}1\\ \phantom{-}0\\  \phantom{-}0\end{pmatrix}$ & $\begin{matrix}\phantom{-}0.82\\ -0.22\\  \phantom{-}0.16\end{matrix}$ & $\begin{matrix}\phantom{-}0.33\\ -0.31\\  \phantom{-}0.16\end{matrix}$ & $\begin{matrix}0.25\\ 0.75\\ 0.18\end{matrix}$
\end{tabular}
}

\caption{World frame coordinates for the impact locations. The number \# identifies the place, the direction vector is given in the world frame and the coordinates are given in the world frame. The origin of the world frame is located at the base frame of the robot.} 
\label{tab:places}
\end{table}

We distinguish three principal cases of experiments:
\begin{enumerate}
    \item no skin -- no protective cover attached to the collision site
    \item passive skin -- protective cover attached to collision site, pressurized but without any collision detection and reaction (on the part of the skin cover)
    \item active skin -- protective cover at collision site with collision detection and connection to robot controller
\end{enumerate}
The last case can be further divided based on the different safety settings for the skin (e.g., Emergency stop and Safeguard stop for UR10e robot). 
Data were separated into individual datasets. Their overview can be found in Tab.\@~\ref{tab:datasets} and they are publicly available at \url{https://osf.io/gwdbm}. Every measurement on the UR10e and KUKA iiwa robots was repeated 3 times. On the KUKA Cybertech, in order to limit mechanical stress to the heavy robot, single measurements were taken. Videos illustrating the experiments are available at \url{https://youtu.be/yqEjnK9_hCg}.

\paragraph{UR10e}
For the UR10e robot, we divided the collected data into three different datasets. The first dataset contains \textit{transient} contact impacts for active (Safeguard stop), passive, and no skin at 9 different velocities (0.2, 0.25, 0.3, 0.35, 0.4, 0.45, 0.5, 0.6, and 0.7~m/s) with the least restrictive safety preset (Pre-4). 
\textit{Quasi-static} contact impacts were collected for active (Safeguard stop), active (Emergency stop), passive, and no skin at 7 different velocities (from 0.2 to 0.5~m/s with 0.05~m/s increment) with the least restrictive safety preset (Pre-4). 
And the third dataset consists of \textit{quasi-static} contact impacts for active (Safeguard stop), passive, and no skin in 5 different velocities (from 0.2 to 0.4~m/s with increment 0.05~m/s) with the second most restrictive safety preset (Pre-2).

\paragraph{KUKA iiwa}
Five datasets were collected with the KUKA iiwa robot. The first dataset contains \textit{transient} contact impacts for active (Stop 0), passive, and no skin in 9 different velocities (0.2, 0.25, 0.3, 0.35, 0.4, 0.45, 0.5, 0.6, and 0.7~m/s), with Stop 0 for external torque limit 10~N. 

The remaining datasets consist of \textit{quasi-static} contact impacts. Three of them contain impacts with all five skin settings combinations --- active skin (Stop 0, Stop 1, Stop 1 op), passive skin and no skin --- at 7 different velocities (from 0.2 to 0.5~m/s with 0.05~m/s increment). Each of these datasets also has a different safety stop (Stop 0, Stop 1, Stop 1 op) setting for external torque limit 10~N. 
The last dataset, with no safety stop for external torque in 7 different velocities (from 0.2 to 0.5~m/s with increment 0.05~m/s), has only active skin safety settings combinations (Stop 0, Stop 1, Stop 1 op). Impacts for the combination with no safety stop for neither external torque, nor skin exceeded 500~N even for low velocities.

\paragraph{UR10e pads comparison}
Since a different AIRSKIN pad for each robot was used, we decided to set up another experiment with the UR10e robot having the impact point covered either by the Pad or UR-skin (experiment ``UR10e pads comparison'' in Tab.\@~\ref{tab:datasets}). In this case, the data are divided into two datasets. The first dataset contains \textit{transient} contact impacts for active Pad or UR-skin (Safeguard stop), passive Pad or UR-skin, and no skin at 3 different velocities (0.2, 0.4, and 0.6~m/s) with the least restrictive safety preset (Pre-4). 
The second one consists of \textit{quasi-static} contact impacts for active Pad or UR-skin (Safeguard stop), passive Pad or UR-skin, and no skin in 3 different velocities (0.2, 0.3, and 0.4~m/s) with the least restrictive safety preset (Pre-4).

\paragraph{KUKA KR Cybertech} 
Since KUKA Cybertech is a non-collaborative robot with a mass of around 250~kg, quasi-static collisions without skin or with passive skin would be dangerous (for the robot, AIRSKIN, and the measuring device). For that reason, we collected a \textit{quasi-static} dataset only for the active skin in both externally triggered stop categories (Stop 1 op, Stop 2) at 5 different velocities (0.2, 0.25, 0.3, 0.35, 0.4~m/s) in directions along $x$-axis and $y$-axis (data along $z$-axis were not collected). 
The \textit{transient} contact impacts were collected for active skin (Stop 1 op) and no skin at 5 different velocities (0.2, 0.3, 0.4, 0.5, 0.6~m/s), because there was no difference in resulting forces between stop categories and active/passive skin for the transient experiments.

\begin{table}[htb]
\centering
\resizebox{250pt}{!}{%
\begin{tabular}{c|c|cc|ccc}
\textbf{\begin{tabular}[c]{@{}c@{}}Robot \& \\ setup\end{tabular}} & \textbf{\begin{tabular}[c]{@{}c@{}}Contact \\ type\end{tabular}} & \textbf{\begin{tabular}[c]{@{}c@{}}Skin \\ type\end{tabular}} & \textbf{\begin{tabular}[c]{@{}c@{}}Skin \\ settings\end{tabular}} & \textbf{\begin{tabular}[c]{@{}c@{}}Safety\\ settings\end{tabular}} & \textbf{\begin{tabular}[c]{@{}c@{}}Velocities\\ {[}m/s{]}\end{tabular}} & \textbf{\begin{tabular}[c]{@{}c@{}}Sam-\\ ples\end{tabular}} \\ \hline
 \multirow{10}{*}{UR10e} & \multirowcell{4}{quasi-\\static} & \multirowcell{3}{UR-\\skin} & S-stop & \multirow{4}{*}{Pre-4} & \multirow{4}{*}{\begin{tabular}[c]{@{}c@{}}0.2, 0.25, 0.3\\ 0.35, 0.4, 0.45\\ 0.5\end{tabular}} & \multirow{4}{*}{252} \\
 &  & & E-stop &  &  &  \\
 &  & & Passive &  &  &  \\ \cline{3-4}
 &  & - & No skin &  &  &  \\ \cline{2-7}
 & \multirowcell{3}{quasi-\\static}  & \multirowcell{2}{UR-\\skin} & S-stop & \multirow{3}{*}{Pre-2} & \multirow{3}{*}{\begin{tabular}[c]{@{}c@{}}0.2, 0.25, 0.3\\ 0.35, 0.4\end{tabular}} & \multirow{3}{*}{135} \\
 &  & & Passive &  &  &  \\ \cline{3-4}
 &  & - & No skin &  &  &  \\ \cline{2-7}
 & \multirow{3}{*}{transient} & \multirowcell{2}{UR-\\skin} & S-stop & \multirow{3}{*}{Pre-4} & \multirow{3}{*}{\begin{tabular}[c]{@{}c@{}}0.2, 0.25, 0.3\\ 0.35, 0.4, 0.45\\ 0.5, 0.6, 0.7\end{tabular}} & \multirow{3}{*}{162} \\
 &  & & Passive &  &  &  \\ \cline{3-4}
 &  & - & No skin &  &  &  \\   \hline
 \multirowcell{21}{KUKA\\iiwa} & \multirowcell{5}{quasi-\\static} & \multirow{4}{*}{Pad} & Stop 0 & \multirow{5}{*}{Stop 0} & \multirow{5}{*}{\begin{tabular}[c]{@{}c@{}}0.2, 0.25, 0.3\\ 0.35, 0.4, 0.45\\ 0.5\end{tabular}} & \multirow{5}{*}{315} \\
 &  & & Stop 1 &  &  &  \\
 &  & & Stop 1 op &  &  &  \\
 &  & & Passive &  &  &  \\ \cline{3-4}
 &  & - & No skin &  &  &  \\ \cline{2-7}
 & \multirowcell{5}{quasi-\\static} & \multirow{4}{*}{Pad}  & Stop 0 & \multirow{5}{*}{Stop 1} & \multirow{5}{*}{\begin{tabular}[c]{@{}c@{}}0.2, 0.25, 0.3\\ 0.35, 0.4, 0.45\\ 0.5\end{tabular}} & \multirow{5}{*}{268} \\
 &  & & Stop 1 &  &  &  \\
 &  & & Stop 1 op &  &  &  \\
 &  & & Passive &  &  &  \\ \cline{3-4}
 &  & - & No skin &  &  &  \\ \cline{2-7}
 & \multirowcell{5}{quasi-\\static} & \multirow{4}{*}{Pad} & Stop 0 & \multirow{5}{*}{\begin{tabular}[c]{@{}c@{}}Stop\\ 1 op\end{tabular}} & \multirow{5}{*}{\begin{tabular}[c]{@{}c@{}}0.2, 0.25, 0.3\\ 0.35, 0.4, 0.45\\ 0.5\end{tabular}} & \multirow{5}{*}{293} \\
 &  & & Stop 1 &  &  &  \\
 &  & & Stop 1 op &  &  &  \\
 &  & & Passive &  &  &  \\ \cline{3-4}
 &  & - & No &  &  &  \\ \cline{2-7}
 & \multirowcell{3}{quasi-\\static} & \multirow{3}{*}{Pad} & Stop 0 &  \multirow{3}{*}{Off} & \multirow{3}{*}{\begin{tabular}[c]{@{}c@{}}0.2, 0.25, 0.3\\ 0.35, 0.4, 0.45\\ 0.5\end{tabular}} & \multirow{3}{*}{173} \\
 &  & & Stop 1 &  &  &  \\
 &  & & Stop 1 op &  &  &  \\ \cline{2-7}
 & \multirow{3}{*}{transient} & \multirow{2}{*}{Pad} & Stop 0 & \multirow{3}{*}{Stop 0} & \multirow{3}{*}{\begin{tabular}[c]{@{}c@{}}0.2, 0.25, 0.3\\ 0.35, 0.4, 0.45\\ 0.5, 0.6, 0.7\end{tabular}} & \multirow{3}{*}{162} \\
 &  & & Passive &  &  &  \\ \cline{3-4}
 &  & - & No skin &  &  &  \\ \hline
 \multirowcell{10}{UR10e\\ Pads\\ compa-\\rison} & \multirowcell{5}{quasi-\\static}  & \multirowcell{2}{UR-\\skin}  & S-stop & \multirow{5}{*}{Pre-4} & \multirow{5}{*}{\begin{tabular}[c]{@{}c@{}}0.2, 0.3, 0.4\end{tabular}} & \multirow{5}{*}{135} \\
 &  & & Passive &  &  &  \\ \cline{3-4}
 &  & \multirow{2}{*}{Pad} & S-stop &  &  &  \\
 &  & & Passive &  &  &  \\ \cline{3-4}
 &  & - & No skin &  &  &  \\ \cline{2-7}
 & \multirow{5}{*}{transient}  & \multirowcell{2}{UR-\\skin}  & S-stop & \multirow{5}{*}{Pre-4} & \multirow{5}{*}{\begin{tabular}[c]{@{}c@{}}0.2, 0.4, 0.6\end{tabular}} & \multirow{5}{*}{90} \\
 &  & & Passive &  &  &  \\ \cline{3-4}
 &  & \multirow{2}{*}{Pad} & S-stop &  &  &  \\
 &  & & Passive &  &  &  \\ \cline{3-4}
 &  & - & No skin &  &  &  \\ \hline
  \multirowcell{4}{KUKA\\Cybertech} & \multirowcell{2}{quasi-\\static} & \multirow{2}{*}{Pad} & Stop 1 op & \multirow{2}{*}{-----} & \multirow{2}{*}{\begin{tabular}[c]{@{}c@{}}0.2, 0.25, 0.3\\ 0.35, 0.4\end{tabular}} & \multirow{2}{*}{20} \\
 &  & & Stop 2 &  &  &  \\ \cline{2-7}
 & \multirow{2}{*}{transient} & Pad & Stop 1 op & \multirow{2}{*}{-----} & \multirow{2}{*}{\begin{tabular}[c]{@{}c@{}}0.2, 0.3, 0.4,\\ 0.5, 0.6\end{tabular}} & \multirow{2}{*}{20} \\ \cline{3-4}
 &  & - & No skin &  &  &  
\end{tabular}
}
\caption{Datasets overview. For details see Sec.~\ref{subsec:stop_cats}, \ref{subsec:airskin}, and \ref{subsec:setup}.}
\label{tab:datasets}
\end{table}

\section{Results}
\label{sec:results}
The presented results consist of four separate parts. First, we present the post-collision behaviors. This is followed by the impact force measurements of the studied robots (UR10e, KUKA iiwa, and Cybertech) in various combinations of AIRSKIN and robot stop settings (see Tab.\@~\ref{tab:datasets}). Then, the effects of the stopping behavior on the impact force are presented. Last, we present the measurements of the AIRSKIN's properties, stiffness and activation threshold force, to demonstrate the possible effect of the skin's properties. A video illustrating a selection of the experiments is available at \url{https://youtu.be/yqEjnK9_hCg}.

\subsection{Post-collision behavior}

\begin{figure}[htb]
	\centering
	\includegraphics[width=0.45\textwidth]{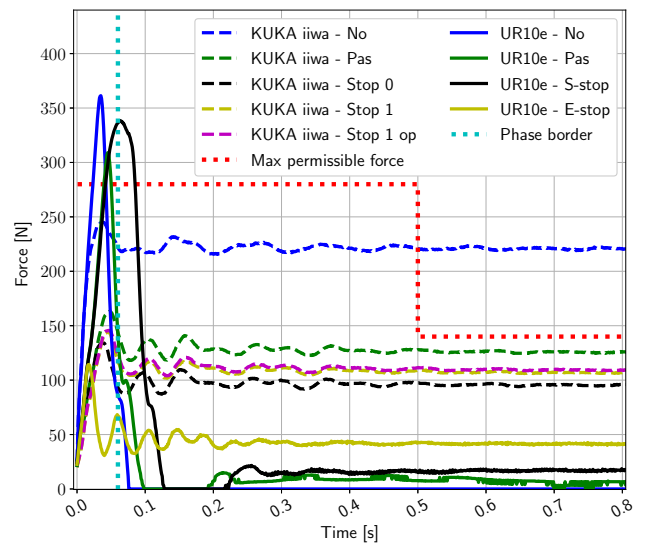}
	\caption{Force evolution after impact for the velocity 0.3~m/s at Place 2. UR10e -- solid lines. KUKA iiwa -- dashed lines. The AIRSKIN module is either absent (No), active (Stop 0/Stop 1/Stop 1 op/S-stop/E-stop) or merely pressurized but not active (Pas) (see Tab.\@~\ref{tab:stops}). An example of an actual phase I / impact phase boundary (cf. Fig~\ref{fig:phases}) , namely for KUKA - skin No --- cyan dotted vertical line. Permissible force per TS~15066 --- red dotted horizontal line. }
	\label{fig:robots_force_comp}
\end{figure}

The robots, due to their proprietary controllers, present different reaction behavior (see also~\cite{Svarny2021}). Upon quasi-static impact, the UR10e generally bounced back, while the KUKA robots stayed at the impact position. This means the impact duration needs to be interpreted differently between the robots and even between collisions. A Safeguard stop based UR10e impact is clearly delimited in the measurements as the robot stops exerting force upon the measuring device. The KUKA iiwa, however, shows a prolonged damped harmonic movement upon impact. See Fig.\@~\ref{fig:robots_force_comp} for comparison between the collaborative robots. In addition, we observed that the Emergency stop of the UR10e also leads to this damped harmonic movement. The probable cause of this movement is the oscillation of the robot and its joint motors after the abrupt stoppage. In these cases, we delimited the impact duration by the measurement onset and the first minimum after the impact peak. The Cybertech collision force profile did not allow the distinction of this first minimum, as the robot continued to push against the measuring device (see Fig.\@~\ref{fig:ct_force}). Therefore some measurements (e.g., phase I collision duration) cannot be presented. However, let us add that the use of AIRSKIN changes the magnitude of the forces and not the shape of the force profile. The force profile depends on the robot and its respective stopping behaviors (see the differences in Fig.\@~\ref{fig:robots_force_comp}).

\subsection{Impact force measurements}
\label{subsec:force_measurements}

We present the collected impact force measurements hereafter. The three robots employed differ in their mechanical properties (mass, degrees of freedom etc.), controllers, and safety settings. Hence, the results across the robots should not be directly compared. Therefore we focused on the comparison between different settings for a given robot (active vs. passive vs. no skin, safety settings, robot presets). This allowed us to look at common trends between the robots.

\subsubsection{UR10e}
\begin{figure*}[!htpb]
\vskip-1em
\centering
\includegraphics[width=0.925\textwidth]{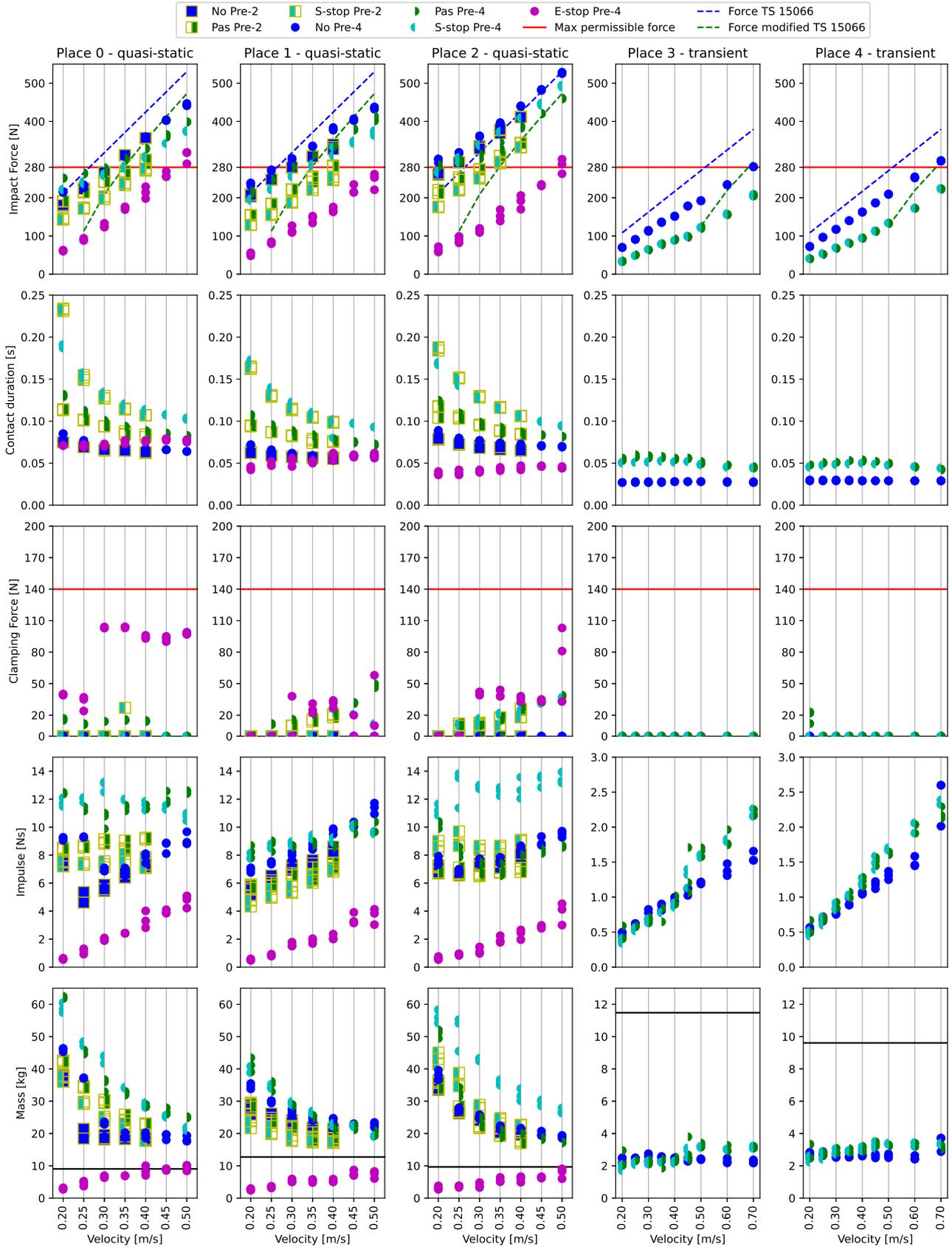}\vskip-1em
\caption{UR10e summary of measurements. The AIRSKIN module is either absent (No), active (S-stop/E-stop) or merely pressurized but not active (Pas). The robot is either in the least restrictive preset (Pre-4) or in the second most restrictive preset (Pre-2). In addition, active skin could trigger either the Safeguard stop (S-stop) or the Emergency stop (E-stop). The duration is the time between the impact detection and the first local minimum of the force measurement, i.e., the end of Phase I. The maximum permissible force is derived from TS~15066. Dashed lines are force predictions from Eq.~\ref{eq:f_pfl_soft} -- see Sec.~\ref{subsec:model_soft}. The effective mass is calculated from the UR10e model.}
\label{fig:ur_summary}
\end{figure*}

We present the results in \fig{\ref{fig:ur_summary}}. The figure can be investigated row by row as each one is a different perspective on the collision. A distinction is visible between the behavior of the first three columns representing quasi-static collisions and the last two columns, the transient cases.

The impact forces grow with the velocity in a similar manner independent of the applied protective measures. However, it is visible that the use of the strict Emergency stop with Pre-4 (skin E-stop), leads to lower impact forces than any other method. It also resulted in impact forces in general below the allowed force threshold. Other settings are not so clearly delimited. For example, the active use of the skin (markers filled only on the left side) generally yielded lower impact forces than passive skin (markers filled only on the right). However, this is not true for Place 2 and Pre-4, where the active skin (S-stop) has higher impact force values than passive skin (Pas). Nevertheless, the first row shows that active skin leads consistently to lower impact forces than when there was no skin (No Pre-2 and No Pre-4). In transient collisions, passive skin leads to the same impact force as active skin; see the matching semicircles in the figure. 

The dashed lines illustrate force predictions with Eq.~\ref{eq:f_pfl_soft} using the standard version from TS 15066 (Sec.~\ref{subsec:nature}) (blue) where no compliant cover is considered and using the modification taking the stiffness and compressible thickness of the cover into account when AIRSKIN is used (green). The modified model prediction (available only for bigger velocities; see Sec.~\ref{subsec:model_soft} for details) indeed better estimates the impact forces when protective skin is employed.

Contact duration in the quasi-static cases shows predominantly three trends. First, for the no skin or E-stop case, the duration is independent of the velocity and almost constant. Second, for S-stop, the contact duration gets shorter with higher velocity (see S-stop). The last trend is transient contacts. These show a very short contact duration for all the settings with a slightly longer duration when AIRSKIN was used. This is probably due to its softness and deformation upon collision. Especially in the case of the passive skin, the absorption of the collision by the skin delays the collision detection by the robot by delaying the moment when the critical threshold is exceeded. 

The clamping force played a significant role only when E-stop was used. Thus while this strict stopping behavior diminished the impact force, it also lead to clamping because the robot joints were stopped by a path maintaining Stop 1 and could not bounce of. Nevertheless, the clamping forces were safely lower than the maximum permissible clamping force (140~N).

The impulse was overall higher or comparable between the use of the skin and no skin. Lowest impulses were measured again with the use of the E-stop. Depending on the place, the impulse could be increasing with velocity or constant, for places 1, 3, 4, and places 0, 2, respectively. Combined with the previously seen contact duration, we can conclude that the skin extends the contact duration and distributes the impact energy in time. However, the constant impulses in Place 0 and 1 show that the skin presence also increased the overall transmitted energy. 

The estimation of the robot mass significantly exceeded the computed effective mass (even five times for UR10e, Place 0 and 2 for low velocities 60~kg instead of its 12.5~kg). This can lead to a hypothesis that the effective mass itself is not enough to determine the collision force characteristics and that the robot dynamics and controller behavior should also be considered.

These findings are also numerically summarized in \tab{\ref{tab:ur_mean_diff}}. Therefrom we can make additional observations. For quasi-static contact and the same safety preset, the difference between the passive and active use of the skin with the UR10e can be as small as only 4~\% (compare the mean for Pas-Pre-4 of -6~\% with S-stop Pre-4 of -10~\%). The use of Pre-2 leads to lower impact forces in general.

\begin{table}[htb]
\centering
\begin{tabular}{c|ccc|c|cc|c}
\multicolumn{8}{c}{\textbf{UR10e Impact force change to skin No Pre-4 (\%)}}   \\ \hline
\multirow{2}{*}{\textbf{Setup}} &  \multicolumn{4}{c|}{\textbf{Quasi-static}} & \multicolumn{3}{c}{\textbf{Transient}} \\
                 &   P0 &   P1 &   P2 &   Mean &   P3 &   P4 &   Mean \\ \hline
Pas Pre-4      &    1 &   -8 &  -11 &     -6 &  -41 &  -39 &    -40 \\ 
No Pre-2  &   -4 &  -10 &   -8 &     -7 &  -- &  -- &    -- \\ 
 S-stop Pre-4     &   -7 &  -16 &   -7 &    -10 &  -41 &  -38 &    -39 \\ 
 Pas Pre-2 &  -12 &  -29 &  -23 &    -21 &  -- & -- &   -- \\ 
S-stop Pre-2 &  -26 &  -39 &  -30 &    -32 & -- & -- &    -- \\ 
E-stop Pre-4   &  -48 &  -59 &  -62 &    -56 &  -- &  -- &  --
\end{tabular}
\caption{Mean difference of peak impact forces for UR10e compared to skin No Pre-4. Pre-2 stands for the second most restrictive and Pre-4 stands for the least restrictive safety preset. The AIRSKIN can either be absent (No), merely pressurized but not active (Pas) or active (E-stop/S-stop). E-stop stands for the scenarios where AIRSKIN activated the Emergency stop while S-stop means that the Safeguard stop was activated.}
\label{tab:ur_mean_diff}
\end{table}

\begin{figure}[htb]
	\centering
	\includegraphics[width=0.49\textwidth,trim={0.5cm 0cm 1.0cm 0cm},clip]{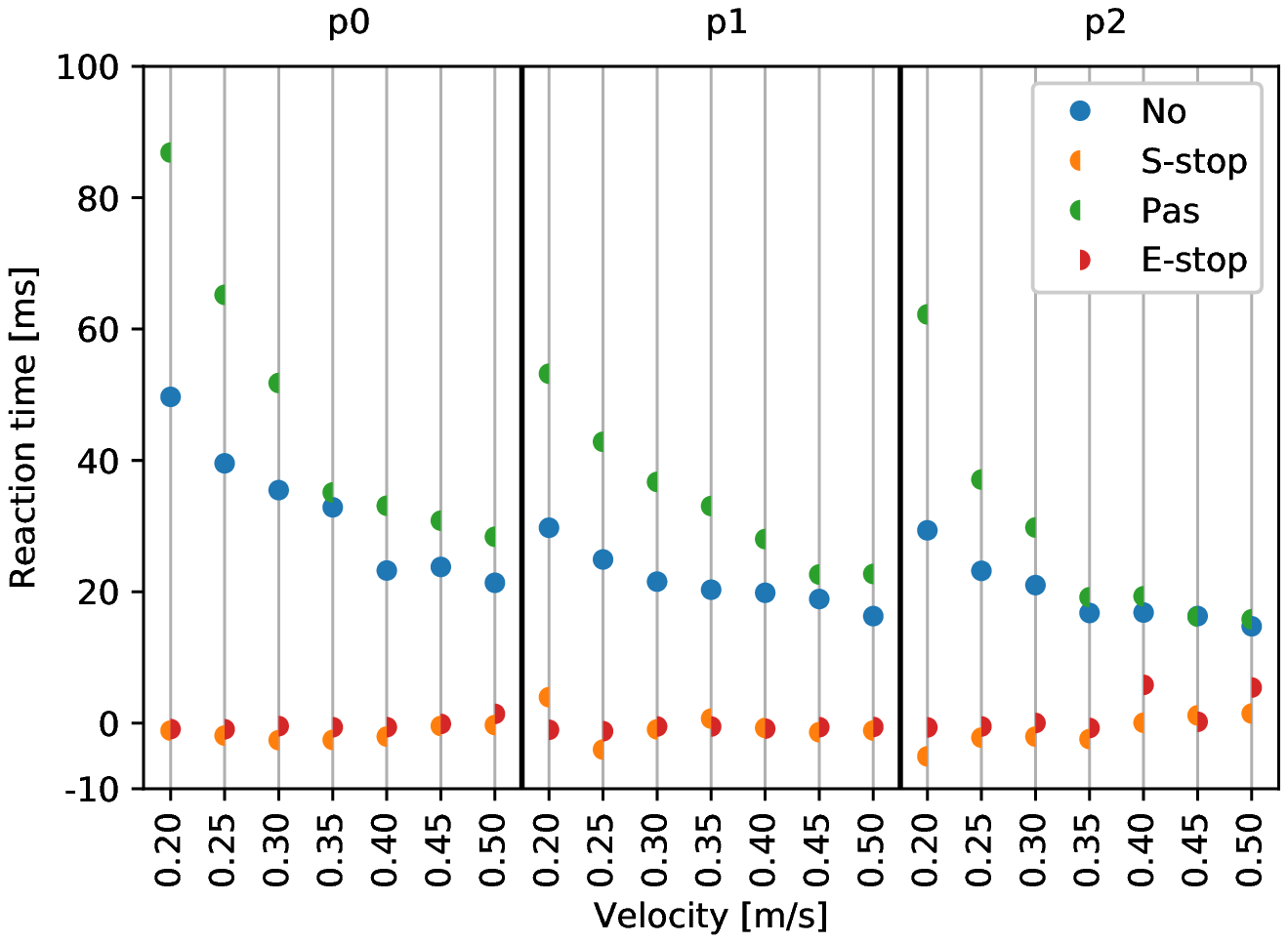}
	\caption{Reaction times for the UR10e using Pre-4. The AIRSKIN can either be absent (No), merely pressurized but not active (Pas) or active (E-stop/S-stop). E-stop stands for the scenarios where the robot used the least restrictive preset (Pre-4) and AIRSKIN activated the Emergency stop instead of the Safeguard stop.}
	\label{fig:ur_reaction_times}
\end{figure}

The reaction times studied on the UR10e are presented in Fig.\@~\ref{fig:ur_reaction_times} and \ref{fig:ur_reaction_times_pad}. If AIRSKIN is allowed to trigger any kind of stop behaviors, then these are initiated at the contact with the skin and before a contact with the robot itself. However, if we rely only on the robot's sensors (i.e., AIRSKIN is passive or not installed), the stopping behavior is initialized only after these sensors detect the collision.

The comparison between Fig.\@~\ref{fig:ur_reaction_times} and \ref{fig:ur_reaction_times_pad} also shows that the passive properties of the pad affect differently the reaction times based on the impact place. The active pad and UR-skin lead to the same reaction times in all the places except for Place 2.

\subsubsection{KUKA iiwa}

\begin{figure*}[htpb]
\vskip-1em
\centering
\includegraphics[width=0.93\textwidth]{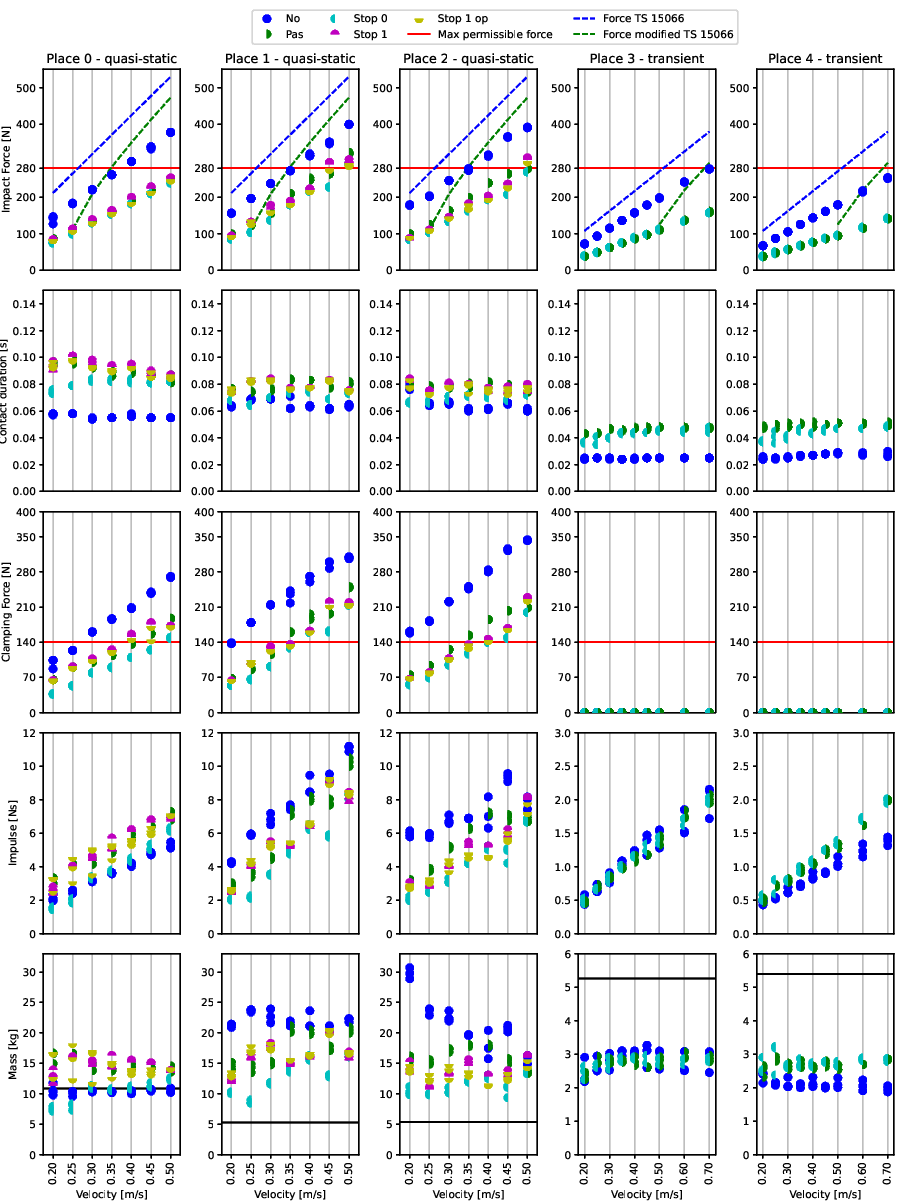} \vskip-1em
\caption{KUKA iiwa summary of measurements. The AIRSKIN module is either absent (No), merely pressurized but not active (Pas) or active with various stopping behaviors (Stop 0, Stop 1, Stop 1 op). The maximum permissible force is derived from TS~15066. Dashed lines are force predictions from Eq.~\ref{eq:f_pfl_soft} -- see Sec.~\ref{subsec:model_soft}. The duration is the time between the impact detection and the first local minimum of the force measurement, i.e., the end of Phase I. The effective mass is calculated from the KUKA iiwa model.}
\label{fig:kuka_summary}
\end{figure*}

Similarly to the UR robot, \fig{\ref{fig:kuka_summary}} summarizes the measurements with respect to five perspectives. The use of the Pad (passive or active) leads to lower impact forces compared to the impacts without the Pad (skin No). These effects are visible in both types of collisions, the quasi-static and transient, and in general lead to lower impact forces than the permissible force. Similarly to the UR10e, the transient collisions result in the same impact forces for both the active and passive Pad shown by the overlap of the semicircles, suggesting that only the passive properties affect the impact force. However, unlike the UR10e, quasi-static collisions lead to clamping forces (see third row in \fig{\ref{fig:kuka_summary}}) exceeding the maximum permissible force of 140~N even with the use of the Pad. 
The dashed lines---force predictions using  Eq.~\ref{eq:f_pfl_soft}---are consistent in the sense that the modified model (green lines) predicts lower collision forces for the elastic cover.

The collision durations are independent of the velocity and are constant. In addition, collision durations for skin No and other cases are small. 
As the duration stayed constant but the impact forces increased with the velocity, the impulse is also increasing. Interestingly, while impact forces for skin No were higher than for the other cases, the impulse is the same or lower.

The KUKA iiwa's mass measurements show the significance of the impact location very clearly. While Place 0 matches the robot's predicted effective mass, in the other two places the resulting effective mass exceeds significantly the predicted value. This is visible especially in the case of skin No.  

The KUKA iiwa data present two different observations for quasi-static and transient cases in Tab.\@~\ref{tab:kuka_mean_diff}, top row. Both cases show the great influence of the passive properties of the Pad (a decrease of 32~\% or even 46~\%). While in the transient collisions the activation of the skin did not show any effect, it can decrease the mean impact force by another 6~\% in the quasi-static case. There is no significant difference between the two variants of Stop 1.

The effect of AIRSKIN is more prevalent if the external torques activate only Stop 1 -- see Tab.\@~\ref{tab:kuka_mean_diff} bottom row. We see a lower effect of merely passive skin (average 26~\% as opposed to 32~\% with ext. torque on Stop 0) but a much larger improvement if the Pad activates a Stop 0 (average 67~\% as opposed to 40~\% with ext. torque on Stop 0).

\begin{table}[htbp]
\centering
\begin{tabular}{c|ccc|c|cc|c}
\multicolumn{8}{c}{\textbf{KUKA iiwa - Impact force change to skin No (\%)}}   \\  \multicolumn{8}{c}{} \\
\multicolumn{8}{c}{\textbf{External torque limit on Stop 0}} \\ \hline
\multirow{2}{*}{\textbf{Setup}} &  \multicolumn{4}{c|}{\textbf{Quasi-static}} & \multicolumn{3}{c}{\textbf{Transient}} \\
                    &   P0 &   P1 &   P2 &   Mean &   P3 &   P4 &   Mean \\ \hline
 Pas        &  -37 &  -28 &  -32 &    -32 &  -45 &  -46 &    -46 \\ 
 Stop 1    &  -36 &  -29 &  -38 &    -35 &  -- &  -- &    -- \\ 
 Stop 1 op &  -38 &  -31 &  -38 &    -36 &  -- &  -- &    -- \\ 
 Stop 0    &  -41 &  -37 &  -43 &    -40 &  -45 &  -45 &    -45 \\ \hline 
 \multicolumn{8}{c}{} \\
\multicolumn{8}{c}{\textbf{External torque limit on Stop 1}} \\ \hline
 Pas           &  -41 &  -14 &  -23 &    -26 &  -- &  -- &    -- \\
 Stop 1    &  -48 &  -20 &  -33 &    -34 &  -- &  -- &    -- \\
 Stop 1 op &  -47 &  -22 &  -35 &    -35 &  -- &  -- &    -- \\
 Stop 0    &  -66 &  -68 &  -67 &    -67 &  -- &  -- &    -- \\
\hline

\end{tabular}
\caption{Mean difference of peak impact forces for KUKA iiwa. The baseline impact force is the robot without any AIRSKIN modules. This value is compared with measurements where the skin is merely pressurized but not active (Pas) or active with various stopping behaviors (Stop 0, 1, 1 op).}
\label{tab:kuka_mean_diff}
\end{table}

\subsubsection{KUKA Cybertech}

\begin{figure}[htb]
	\centering
	\includegraphics[width=0.49\textwidth]{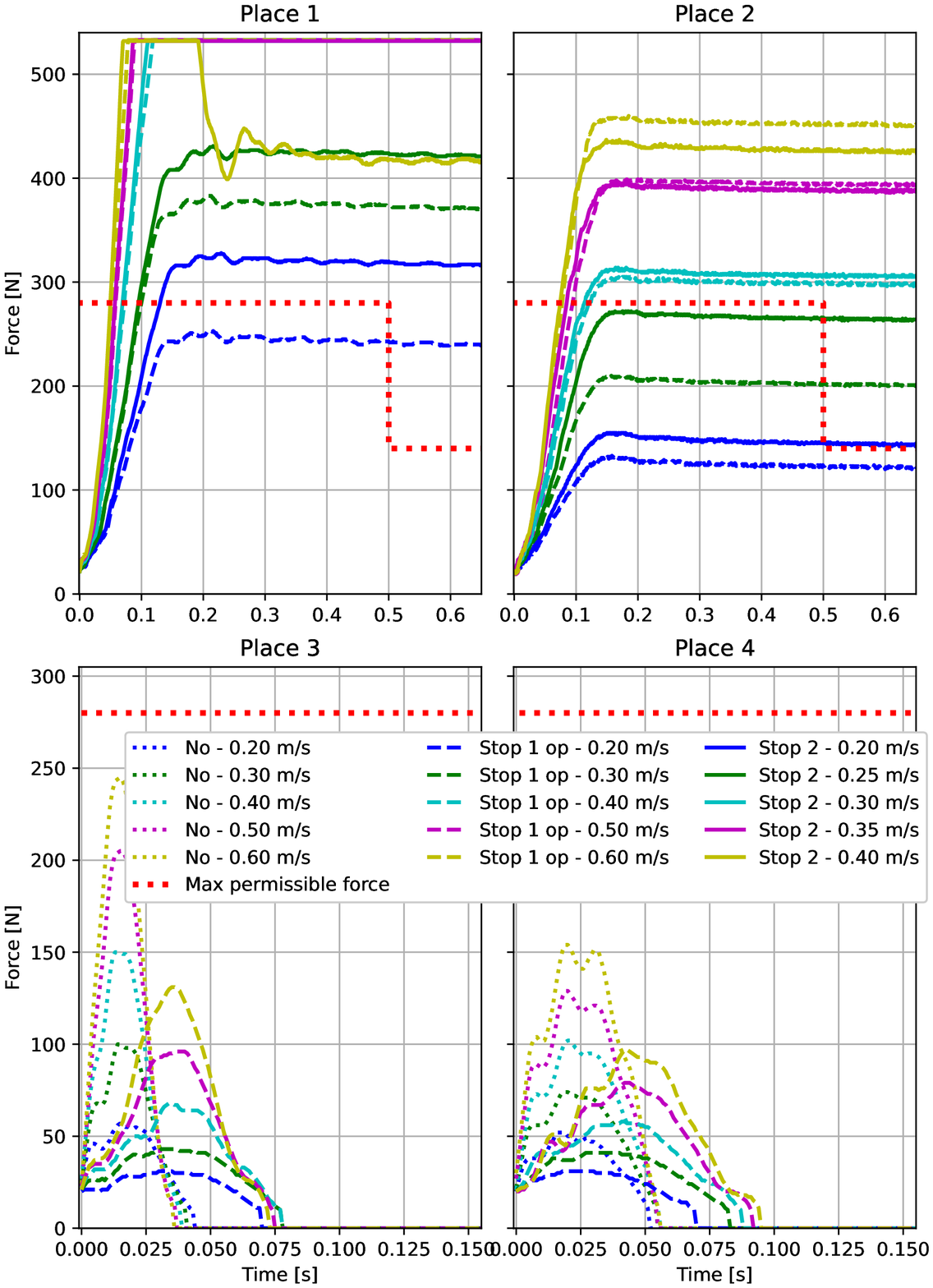}
	\caption{Time courses of measured force for KUKA Cybertech quasi-static (top) and transient (bottom) experiments for the $[0, 1, 0]$ and $[1, 0, 0]$ direction. No skin (No) or with the active Pad initializing a specific stop category (Stop 1 op or Stop 2).
	}
	\label{fig:ct_force}
\end{figure}

Cybertech is an industrial robot with a significantly higher weight than collaborative robots and therefore results in different force profiles as shown in \fig{\ref{fig:ct_force}}. The results present two diametrically different outcomes for quasi-static and transient collisions. Quasi-static collisions, the top row of \fig{\ref{fig:ct_force}}, lead to clamping behavior without significant oscillation (denoted Type 3 in Sec.\@~\ref{subsec:nature}). Measurements could not be performed at Place 0 as the mounting of the pads on this robot did not permit to hit the measuring device with the downward movement. Velocities above 0.30~m/s lead to impact forces above 500~N at Place 1 too. Therefore we could not collect data to study trends as with the previous robots and focus only on the force profiles. Additionally, a special case is the collision at 0.40~m/s for Stop 2 where the impact moved the supporting table. The collected data show an interesting trend at Place 2. While at low velocities Stop 1 leads to lower forces, for velocities higher than 0.30~m/s, the pattern is switched, and Stop 2 leads to lower impact and clamping forces.

Transient collisions, the bottom row of \fig{\ref{fig:ct_force}}, lead to very short contacts where even the robot without protective skin (No) did not exceed the prescribed limit of 280~N. The use of the skin, nevertheless, significantly lowers the resulting impact force. A similar decrease as with the active skin can be achieved if Place 4 is used instead of Place 3. Also visible from the data is the prolonging effect of the soft protective cover when a collision with active skin can lead to a measurable force during a period almost twice as long as a collision without the skin (see Place 4).

Therefore, based on the transient collision data, Cybertech could be used in collaborative scenarios even without an active safety skin if it would be certain that only transient collisions without any clamping occur. However, any risk of quasi-static collisions, even for the low velocities, requires the usage of a device like AIRSKIN.

\subsubsection{Measured forces summary}

Based on the results from particular robots presented above, we can summarize that the application of the skin leads to lower impact forces and the use of stricter stop categories leads to lower peak impact forces. In addition, we can observe an agreement across robots that in the transient collisions there is no difference in peak impact forces between active and passive skins (see \tab{\ref{tab:ur_mean_diff}}, \tab{\ref{tab:kuka_mean_diff}}, \fig{\ref{fig:ct_force}}). However, there are also differences between the robots as prominently visible on the different force profiles already presented in \fig{\ref{fig:phases}}.

\subsection{Stopping behavior effect on the impact forces}
\label{subsec:stop_effect}

\begin{figure*}[htb]
\centering
\includegraphics[width=\textwidth]{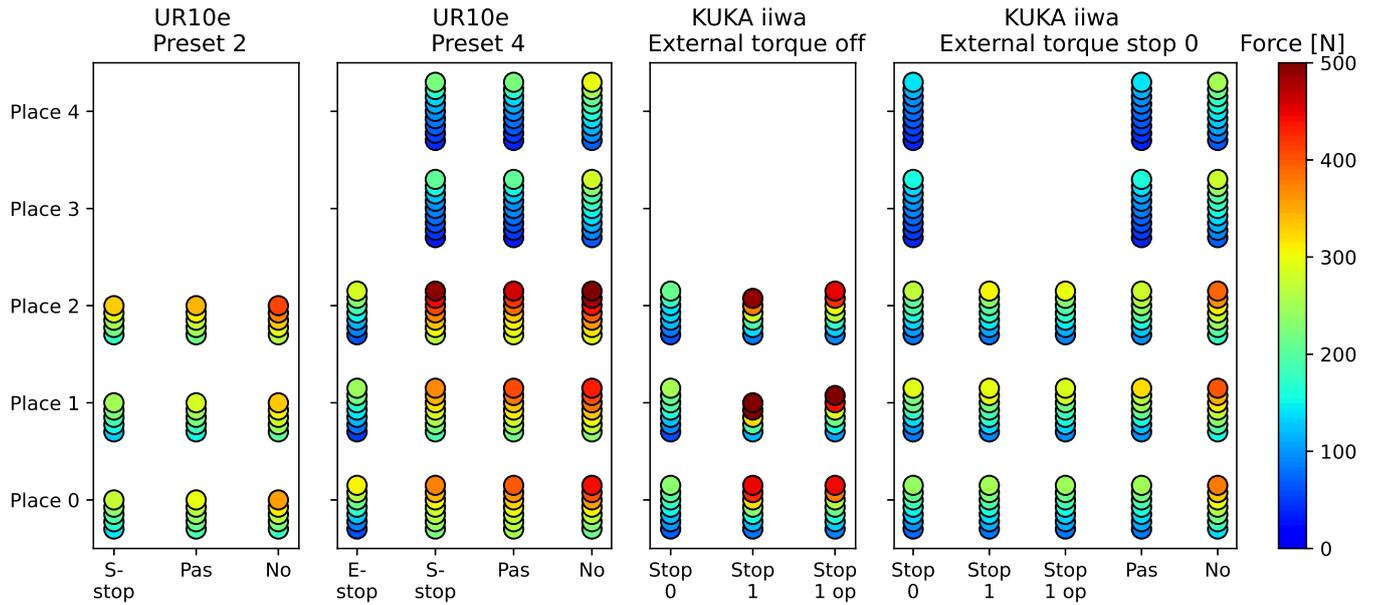} 
\caption{Peak impact forces comparison for the various UR10e and KUKA iiwa with various external torque settings. The circles represent the measured 7 velocities (from 0.2 to 0.5~m/s with increment 0.05~m/s), where applicable. The initiated stop behavior is either a Stop 0, Stop 1, or Stop 1 op on KUKA iiwa. For UR10e, these stops were Safeguard stop (S-stop) and Emergency stop (E-stop) and also the safety preset was considered (Pre-2 or Pre-4). The `Pas' in the case of the AIRSKIN means the pads are pressurized but they do not initiate a stop, while `No' means the AIRSKIN pad was removed from the robot. The three locations Place 0, 1, 2 are quasi-static collisions in the three directions downward, along $y$-axis, along $x$-axis respectively. The transient collisions along $y$-axis and along $x$-axis are Place 3, 4 respectively.}
\label{fig:2d_comparison}
\end{figure*}

The effects of the stopping behavior settings for the collaborative robots are summarized in \fig{\ref{fig:2d_comparison}}. The results are separated based on the robot-specific safety setting (Pre-2 or Pre-4 for UR10e and using the external torque limit for the KUKA iiwa). The horizontal axis then captures the various stops triggered by AIRSKIN if it is present or the skin can be passive (`Pas') or absent (`No'). The measured peak impact forces are then represented with separate circles for each velocity increment (from 0.2 to 0.5~m/s with increment 0.05~m/s). The vertical axis shows the various impact locations (Place 0--4).

The UR10e measurements in Fig.\@~\ref{fig:2d_comparison} and \tab{\ref{tab:ur_mean_diff}} also showed a significant effect of the specific stopping behavior of the robot. The skin improved the safety of the operation most when it was combined with the strictest stopping action or safety preset (Pre-2 in our case). This effect was smaller with the KUKA iiwa (\tab{\ref{tab:kuka_mean_diff}}) if the external torque limit was active. However, without the external torque limit, the stopping behavior triggered by the skin became necessary to stop the robot.

\begin{figure*}[htb]
\includegraphics[width=\textwidth]{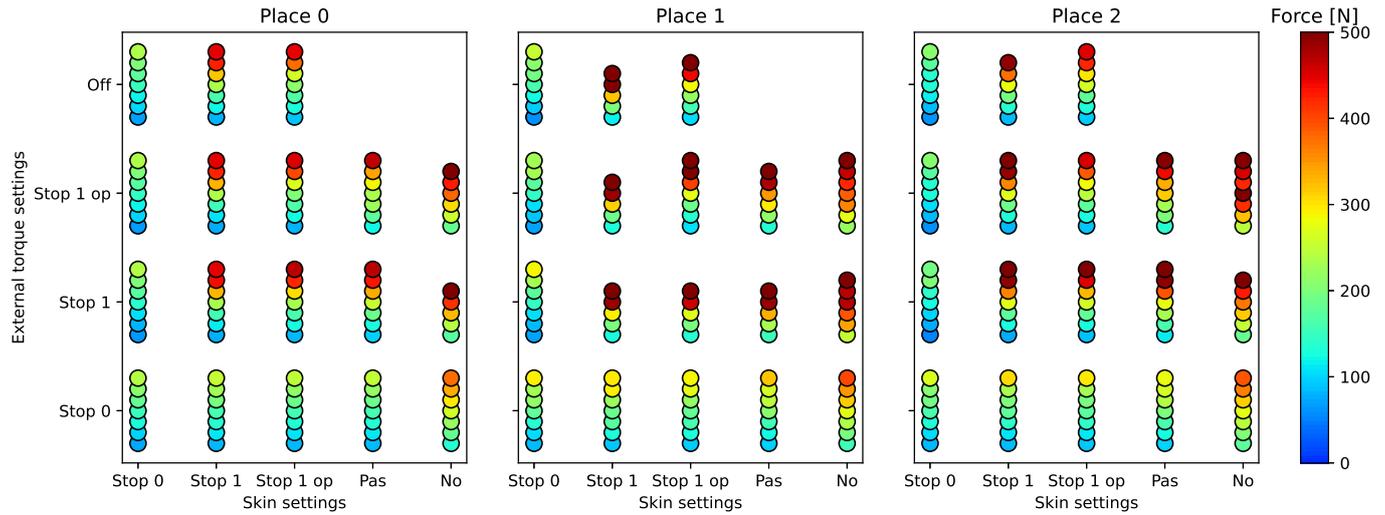}
\caption{KUKA iiwa various stop combinations for quasi-static impacts. The initiated stop behavior is either a category 0 stop (Stop 0), a category 1 stop (Stop 1), a category 1 stop on path (Stop 1 op). The `Pas' in the case of the AIRSKIN means the pads are pressurized but they do not initiate a stop, while `No' means the AIRSKIN pad was removed from the robot. The `Off' setting for the torques means that they were turned off. The circles represent the measured 7 velocities (from 0.2 to 0.5~m/s with increment 0.05~m/s), where applicable. The three locations Place 0, 1, 2 are downward, along $y$-axis, along $x$-axis respectively.}
\label{fig:kuka_stops}
\end{figure*}

Since we did not have the possibility to control all the stop categories for the UR10e, we investigated the effects of the various stopping behaviors only with the KUKA iiwa robot.
All the KUKA iiwa quasi-static impact measurements are presented in Fig.\@~\ref{fig:kuka_stops} to demonstrate the effect of the various stopping behaviors on the final impact force in addition to \fig{\ref{fig:2d_comparison}}. They are organized by the velocities into up to seven dots. In some cases, the exerted forces were higher than the measuring limit of our device (500~N) and thus we did not continue measuring for higher velocities or less safe setups (e.g., the `noskin' columns). The KUKA iiwa robot allowed us to compare its external-torque-based stopping behavior with the AIRSKIN pad-based reaction between all three stop categories. The results support the earlier observation that AIRSKIN combined with a restrictive stop provides the best benefits.

Notice the first column where AIRSKIN triggers a Stop 0. It shows that the impact force stays low for all the measured velocities and even if the external torque trigger is not used. This finding was consistent across all the locations, as also visible in \fig{\ref{fig:2d_comparison}}. According to our data, AIRSKIN can serve as a replacement of external torque sensing. This is visible also in case of the Cybertech robot, see Fig.\@~\ref{fig:ct_force}, where the more restrictive Stop 1 leads to lower impact forces.

Next to events generated by the active AIRSKIN, there may be safety events triggered by the collaborative robot itself. On the UR10e robot, collisions detected by the robot are handled internally by the robot controller. However, on the KUKA iiwa, the user can define how external torque limit violations are processed. Namely, one can choose whether this is connected to Stop 0, Stop 1, or Stop 1 op. This is visible in \fig{\ref{fig:kuka_stops}}, where the peak impact forces for the three quasi-static impact places are presented with a combination of stopping behaviors. The settings of the skin are separate (horizontal axis). The skin would either be active and trigger the various stops, or be merely pressurized but passive (Pas), or it would not be present at all (No). The vertical axis captures the stops triggered by the KUKA external torque sensing capability or whether it was turned off (`Off'). The measured 7 velocities (from 0.2 to 0.5~m/s with increment 0.05~m/s) are shown as circles. Therefore, for example, the figure shows that if the skin triggers a Stop 0, the external torques have little effect on the impact forces (see the first column). However, this is not true for the inverse (see bottom row), as an external torque triggered Stop 0 can still lead to impact forces close to 400~N if the robot is not equipped with the Pad. However, if the Pad is used then the activity of the skin has little effect on the resulting forces. 

The upper right corners in \fig{\ref{fig:kuka_stops}} do not contain measurements as with the given settings, the impact forces exceeded the 500~N limit of the measuring device. It is also visible that the impact places influence the resulting forces as at Place 1, the limit of 500~N was exceeded also with other stop combinations (see second column for Stop 1).

However, we can also notice the importance of the collision location from Fig.\@~\ref{fig:kuka_stops}, namely the collisions at Place 1 (along the $y$-axis). While the general observations made so far still hold, we can notice that the exerted forces are larger with AIRSKIN. Similarly the impact places show an effect in Fig.\@~\ref{fig:2d_comparison} where Place 2, in general, registers higher peak impact forces.

\subsection{AIRSKIN module properties}
\label{subsec:airskin_properties}
\begin{figure}[htb]
	\centering
	\includegraphics[width=0.5\textwidth]{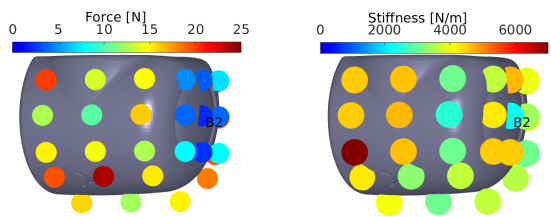}
	\vspace*{-3em}
	\caption{UR-skin threshold force (left) and stiffness (right). The measured values are color-coded. The impact point, B2, is also marked.}
	\label{fig:ur_stif}

\end{figure}

\begin{figure}[htb]
	\centering
	\includegraphics[width=0.5\textwidth]{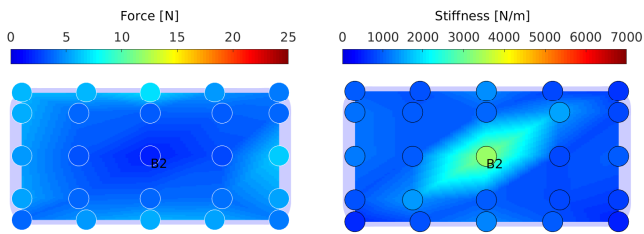}
	\vspace*{-2em}
	\caption{The Pad threshold force (left) and stiffness (right). The measured values are color-coded. The impact point, B2, is also marked.}
	\label{fig:kuka_stif}
\end{figure}

In this work, two types of AIRSKIN modules were employed: UR-skin (used on the UR10e robot, Fig.\@~\ref{fig:ur_pad}) and Pad --- AIRSKIN module pads employed on the KUKA robots (Fig.\@~\ref{fig:kuka_pad}). To assess to what extent these two versions affect the results, we studied two key properties: activation force threshold and mechanical stiffness. The point B2 marks the center of the impact (see Fig.~\ref{fig:ur_stif} and Fig.~\ref{fig:kuka_stif}). 

The threshold force, i.e., the force at which AIRSKIN detects a collision, was 2~N for the UR-skin at point B2 and up to 8~N in the surrounding area (\fig{\ref{fig:ur_stif}}, left). The threshold force for the Pad was less than 2~N at point B2 and less than 4~N in the surrounding area (\fig{\ref{fig:kuka_stif}}, left). Thus, in terms of threshold force, the two locations are comparable. In terms of stiffness, again, the measured values at location B2 are comparable (Fig.~\ref{fig:ur_stif} and Fig.~\ref{fig:kuka_stif} right). 
Overall, UR-skin features much bigger differences in both properties across its surface, while the Pad is more uniform, with the exception of the central area that is stiffer (as there are electronics underneath).

\begin{figure}[htb]
	\centering
	\includegraphics[width=0.48\textwidth]{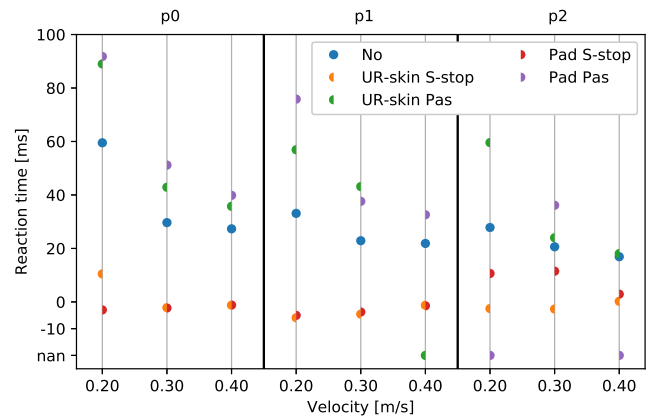}
	\caption{Reaction times for the UR10e with Pad and UR-skin. Reaction times are shown on the vertical axis while collision velocities are on the horizontal axis. The UR10e is either used without any skin or it uses UR-skin or the Pad in a passive regime (UR-skin Pas, Pad Pas) or the skin activates the Safeguard stop (UR-skin S-stop, Pad S-stop).}
	\label{fig:ur_reaction_times_pad}
\end{figure}

Impact force measurements with both types of AIRSKIN on UR10e (Tab.\@~\ref{tab:ur_pad_mean_diff}) show clearly lower average peak impact forces for the Pad compared to UR-skin, although the reaction times are longer (Fig.\@~\ref{fig:ur_reaction_times_pad}).

While the active use of both the skin and the pad outperform their passive use in in the majority of quasi-static situations, this is not true for Place 2 where the opposite holds. The fast reaction time of the active AIRSKIN modules plays an important role in the overall performance as the UR10e starts to brake as soon a contact occurs with the skin (Fig.\@~\ref{fig:ur_reaction_times}). This reaction time explains also the lack of difference between the transient impact forces for the passive and active variants.

However, note that even a passive Pad performs better than the active UR-skin. Therefore the material used for the Pad plays a crucial role.

\begin{table}[htb]
\centering
\resizebox{250pt}{!}{%
\begin{tabular}{cc|ccc|c|cc|c}
\multicolumn{9}{c}{\textbf{UR10e impact force change to skin No (\%)}}  \\ \hline
\multicolumn{2}{c|}{\multirow{2}{*}{\textbf{Setup}}} &  \multicolumn{4}{c|}{\textbf{Quasi-static}} & \multicolumn{3}{c}{\textbf{Transient}} \\
      &    & P0   & P1   & P2   & Mean   &  P3  &  P4  &  Mean  \\ \hline
\multicolumn{1}{c|} {\multirowcell{2}{UR-\\skin}} & Pas &    4 &   -6 &  -10 &     -4 &  -41 &  -38 &    -39 \\
 \multicolumn{1}{c|}{} & S-stop &   -4 &  -16 &   -6 &     -9 &  -40 &  -38 &    -39 \\ \hline
 \multicolumn{1}{c|}{\multirowcell{2}{Pad}} & Pas  &   -5 &  -18 &  -26 &    -16 &  -45 &  -53 &    -49 \\ 
 \multicolumn{1}{c|}{} & S-stop  &  -13 &  -27 &  -21 &    -20 &  -46 &  -54 &    -50 
\end{tabular}
}
\caption{Mean difference of peak impact forces for UR10e with the Pad and the UR-skin. The baseline impact force is the robot without any AIRSKIN pads. The values are then compared to either merely pressurized but not active (Pas) or active skin triggering Safeguard stop (S-stop) modules. The robot always used the Pre-4 preset.}
\label{tab:ur_pad_mean_diff}
\end{table}

\subsubsection{AIRSKIN pad type effect on the impact force}
\label{subsubsec:pad_effect}

The measurements also support the importance of the skin's passive properties, as visible in the comparison between the skin No and skin Pas columns in Fig.\@~\ref{fig:2d_comparison}. The passive skin by itself can significantly lower the impact force, by up to 21~\% (if paired with appropriate stopping behavior; see Tab.\@~\ref{tab:ur_mean_diff}). This also supports the importance of the skin's material properties of the used module. As visible in Tab.\@~\ref{tab:ur_pad_mean_diff} and also Fig.\@~\ref{fig:2d_comparison}, the Pad shows a significant improvement (even as large as the difference between having no AIRSKIN and using the URskin) in performance which can be traced back to its properties.

\section{Conclusion and Discussion}
We performed a total of 2250 measurements of impact forces in five scenarios and various velocities using two collaborative robots, UR10e and KUKA LBR iiwa 7 R800, and the industrial robot KUKA Cybertech KR 20 R1810-2, all equipped with AIRSKIN safety covers. The dataset---involving collisions with active and passive protective skin as well as without it---is publicly available at \url{https://osf.io/gwdbm}. The main findings were the following.

First, we discuss transient collisions. Here, all the robots investigated showed similar behaviors. For the velocities used, even the non-collaborative KUKA Cybertech did not exceed impact collision force limits. We found significant effects of the passive properties of the protective covers---pressurized air in this case---on the transient collision. This was practically the only effect of the skin, as the active skin resulted in almost identical collision evolution and impact forces. Concretely, for transient contacts, passive skin covers resulted in 40~\% lower peak impact force for the UR10e, 45~\% lower for KUKA iiwa, and 46~\% lower for KUKA Cybertech. However, with passive skin, the robot continues along its planned trajectory and clamping may eventually occur. With active skin, the robot will stop after contact is detected. 

Second, we discuss quasi-static collisions. As expected, the peak impact forces are overall much higher. Unlike for transient collisions, the rest of the results has to be discussed taking the robot individual safety settings into account. In these cases, the effect of skin covers---active or passive---cannot be isolated from the collision detection and reaction by the robot itself.
With passive skin covers, the contact duration is prolonged (200~\% compared to no skin for UR10e; 25--50~\% for the KUKA iiwa). On the UR10e,
the effect on peak collision forces is moderate (decrease of 7~\% for Pre-4 and 15~\% for Pre-2). Still, the passive skin cover would allow to move from the more restrictive preset (Pre-2) to a less restrictive (Pre-4), while keeping the collision forces the same. On the KUKA iiwa, the situation is more complicated as we need to consider the connection of the external torque safety setting (see Table~\ref{tab:kuka_mean_diff} and Fig. \ref{fig:kuka_stops} and compare the columns `Pas' and `No'). The biggest effects of passive skin padding are visible when the external torques are connected to Stop 0. In this case, the impact forces are lower by 32~\% on average. For Stop 1, the effect is smaller (26~\%).
On the KUKA Cybertech, measurements with only passive skin were not possible. 

With quasi-static collisions and active skin protection, the safety settings of the robot play a major part. For the UR10e manipulator, we found dramatic effects of the safety settings, namely to which safety stop category the skin was connected (Table~\ref{tab:ur_mean_diff}). In the recommended setting, S-stop, the decrease in peak impact forces was 10~\% for Pre-4 and 26~\% in Pre-2 compared to no skin, but only 4~\% and 13~\%, respectively, compared to passive skin. However, the effect was much larger when connecting to the E-stop, namely 56~\% compared to no skin and 53~\% compared to passive skin. Note that this cannot be explained by the reaction times of the robot (Fig.~\ref{fig:ur_reaction_times}), as these are similar for both stop types. It is the reaction of the robot that is responsible for the measured differences (cf. also Table~\ref{tab:stops}). For the KUKA iiwa, one needs to consider the combination of safety settings: connections of ext. torque and the active skin (Fig.~\ref{fig:kuka_stops}). First, Stop 0 clearly leads to the lowest impact forces. Second, the effects of external torque and skin safety settings are largely interchangeable. Moreover, they may even interfere with each other (e.g., `Stop 1 ext. torque' with `Stop 1 op skin' leads to higher impact forces than `Stop 1 ext. torque' with `Pas skin' or `ext. torque off' and `Stop 1 op skin'). Third, there seems to be no measurable effect of the active skin compared to passive skin. Thus, for the KUKA iiwa, ext. torque safety settings seem to suffice to warrant safety. However, there may be applications where external torques occur naturally and hence active skin may still be needed for safety.
On the contrary, on the industrial robot, KUKA Cybertech, active skin is the only means to make the manipulator collaborative.

Next to empirical measurements, we also provided an extension of the simple collision model of TS~15066 that allows to consider the stiffness and compressible thickness of the protective cover (Sec.~\ref{subsec:model_soft}). Impact force predictions of this model are more in line with the measured data (Fig.~\ref{fig:ur_summary} and \ref{fig:kuka_summary} --- first row) and this extension may thus be considered in future versions of collaborative robot standards for collisions with compliant surfaces.

Furthermore, we studied the force evolution after impact, schematically illustrated in Fig.~\ref{fig:phases}. The first, unconstrained dynamic impact, was observed in the UR10e robot. After collision is detected, the robot seems to actively retract, preventing clamping. The KUKA Cybertech displayed a constrained dynamic impact with clamping. Finally, the KUKA iiwa behaved similarly, but with an additional oscillation in the force profile. Importantly, the presence of passive or active skin protection did not alter this type of behavior (Fig.~\ref{fig:robots_force_comp}). The only case when the force evolution type changed on the same robot was on the UR10e when the skin was connected to the E-stop. 

Let us compare our empirical results with what the collaborative standard TS~15066~\cite{ISOTS15066} prescribes (Sec.~\ref{subsec:nature}). 
For transient contacts, the maximum force limit is 280~N. In our experiments, this limit was never exceeded for all the velocities (from 0.2 to 0.7~m/s), not even for the non-collaborative KUKA Cybertech, where no collision was detected and no reaction triggered. Regarding quasi-static contacts, this is constituted by the force evolution during the first 0.5~s after impact---where the 280~N force limit applies---and the force evolution after the first 0.5~s, for which half of the force threshold, i.e., 140~N, applies. 

The corresponding maximum permissible velocities for the collaborative robots used here are calculated in Section~\ref{subsec:nature} and presented in the ``TS 15066'' column of Table~\ref{tab:ur10e_isots_comp} (UR10e) and Table~\ref{tab:kuka_isots_comp} (KUKA iiwa). The modified limits that take the stiffness and compressible thickness of the protective cover into account are in the ``mod. TS 15066'' column. Velocities that comply with the force limits computed from our empirical measurements are presented in the subsequent columns of these tables for the different settings. The maximum permissible velocity can be obtained as a minimum of the velocities for the first and second phase after impact.

 Also note that for the UR10e robot (Table~\ref{tab:ur10e_isots_comp}), with the exception of AIRSKIN connected to E-stop, no velocity limits result from the second phase (after 0.5~s). This is because despite the clamping nature of the scenario, no actual clamping occurs as the controller allows the robot bounce back (see Fig.~\ref{fig:robots_force_comp} and Fig.~\ref{fig:phases} --- Type 1). Depending on the place in the workspace and the robot preset, robot velocities of 0.2 to 0.3~m/s can be safely operated. Addition of passive or active skin increases the safe velocity by approximately 0.05 to 0.1~m/s. If AIRSKIN is connected to the E-stop, velocities of 0.45~m/s become possible --- contrasting with the TS~15066 prescription of 0.13~m/s (clamping scenario).

\begin{table}[htb]
\centering
\resizebox{250pt}{!}{%
\begin{tabular}{c|c|c|c|ccc|cccc}
 &  &  & & \multicolumn{3}{c|}{\textbf{UR10e Pre-2}} & \multicolumn{4}{c}{\textbf{UR10e Pre-4}} \\ 
\multirow{-2}{*}{\textbf{Place}} & \multirow{-2}{*}{\textbf{\begin{tabular}[c]{@{}c@{}}T \\ {[}s{]}\end{tabular}}} & \multirow{-2}{*}{\textbf{\begin{tabular}[c]{@{}c@{}}TS\\ 15066\end{tabular}}}  & \multirow{-2}{*}{\textbf{\begin{tabular}[c]{@{}c@{}}mod. TS\\ 15066\end{tabular}}}  & No & Pas & \begin{tabular}[c]{@{}c@{}}S-\\ stop\end{tabular} & No & Pas & \begin{tabular}[c]{@{}c@{}}S-\\ stop\end{tabular} & \begin{tabular}[c]{@{}c@{}}E-\\ stop\end{tabular} \\ \hline
 & \textless{}0.5 & 0.26 & 0.35 & \cellcolor{gray!30}0.3 & \cellcolor{gray!30}0.35 & \cellcolor{gray!30}0.4 & \cellcolor{gray!30}0.3 & \cellcolor{gray!30}0.3 & \cellcolor{gray!30}0.3 & \cellcolor{gray!30}0.45 \\
\multirow{-2}{*}{0} & \textgreater{}0.5 & \cellcolor{gray!30}0.13 &\cellcolor{gray!30}0.26 & -- & -- & -- & -- & -- & -- & \textgreater{}0.5 \\ \hline
 & \textless{}0.5 & 0.26 & 0.35 & \cellcolor{gray!30}0.3 & \cellcolor{gray!30}0.35 & \cellcolor{gray!30}0.4 & \cellcolor{gray!30}0.25 & \cellcolor{gray!30}0.25 & \cellcolor{gray!30}0.3 & \cellcolor{gray!30}0.5 \\
\multirow{-2}{*}{1} & \textgreater{}0.5 & \cellcolor{gray!30}0.13&\cellcolor{gray!30}0.26 & -- & -- & -- & -- & -- & -- & \textgreater{}0.5 \\ \hline
 & \textless{}0.5 & 0.26 & 0.35 & \cellcolor{gray!30}0.3 & \cellcolor{gray!30}0.3 & \cellcolor{gray!30}0.3 & \cellcolor{gray!30}\textless{}0.2 & \cellcolor{gray!30}\textless{}0.2 & \cellcolor{gray!30}0.2 & \cellcolor{gray!30}0.45 \\
\multirow{-2}{*}{2} & \textgreater{}0.5 & \cellcolor{gray!30}0.13&\cellcolor{gray!30}0.26 & -- & -- & -- & -- & -- & -- & \textgreater{}0.5
\end{tabular}
}
\caption{Maximum safe end effector velocities --- UR10e [m/s]. Permissible velocities provided by TS 15066 and mod. TS 15066 taking skin compliance into account (see Eq.~\ref{eq:v_pfl_soft}), and safe velocities determined from empirical measurements that do not exceed the collision force (280~N for $T< 0.5~s$, 140~N for $T > 0.5~s$) `No' --- no skin; `Pas' --- passive skin; `S-stop' / `E-stop' --- active skin and its connection to robot safety inputs. Values in gray are minima of the corresponding rows for first and second phase after impact.}
\label{tab:ur10e_isots_comp}
\end{table}

The situation is more complicated for the KUKA iiwa (Table~\ref{tab:kuka_isots_comp}) due to the optional involvement of the external torque limits and, importantly, because the second phase after the collision is present for this robot (see Fig.~\ref{fig:robots_force_comp} and Fig.~\ref{fig:phases} --- Type 2). When relying on the active skin only (ext. torque off), speeds higher than the norm prescribes (0.16~m/s for this robot and the clamping nature) can be safely operated. Namely, 0.2 to 0.25~m/s in the case of `Stop 1' / `Stop 1 op'. A more significant productivity boost, 0.4~m/s safe velocity, constitutes a `Stop 0' connection of the skin. Adding protective skin on top of ext. torque protection (right side of Table~\ref{tab:kuka_isots_comp}), brings about an increase in safe velocity from approx. 0.2~m/s to 0.3~m/s (passive skin) or 0.35~m/s (active skin). 

Importantly, for both robots, the ``mod. TS 15066'' predictions (Eq.~\ref{eq:v_pfl_soft}) are overall more accurate for the situations when protective skin is used than the original values from TS~15066.

\begin{table}[htb]
\centering
\resizebox{250pt}{!}{%
\begin{tabular}{c|c|c|c|ccc|ccccc}
 \multicolumn{3}{c}{} & \textbf{mod.}  & \multicolumn{3}{c|}{\textbf{\begin{tabular}[c]{@{}c@{}}KUKA iiwa \\ external Torque \\ limit off\end{tabular}}} & \multicolumn{5}{c}{\textbf{\begin{tabular}[c]{@{}c@{}}KUKA iiwa \\ external Torque limit \\ Stop 0\end{tabular}}} \\ 
\multirow{-2}{*}{\textbf{Place}} & \multirow{-2}{*}{\textbf{\begin{tabular}[c]{@{}c@{}}T \\ {[}s{]}\end{tabular}}} & \multirow{-2}{*}{\textbf{\begin{tabular}[c]{@{}c@{}}TS\\15066\end{tabular}}}
& \multirow{-2}{*}{\textbf{\begin{tabular}[c]{@{}c@{}}TS\\15066\end{tabular}}}& \begin{tabular}[c]{@{}c@{}}Stop \\ 0\end{tabular} & \begin{tabular}[c]{@{}c@{}}Stop \\ 1\end{tabular} & \begin{tabular}[c]{@{}c@{}}Stop \\ 1 op\end{tabular} & No & Pas & \begin{tabular}[c]{@{}c@{}}Stop \\ 0\end{tabular} & \begin{tabular}[c]{@{}c@{}}Stop\\ 1\end{tabular} & \begin{tabular}[c]{@{}c@{}}Stop\\ 1 op\end{tabular} \\ \hline
 & \textless{}0.5 & 0.32 & 0.43 & 0.5 & 0.35 & 0.4 & 0.35 & 0.5 & 0.5 & 0.5 & 0.5 \\
\multirow{-2}{*}{0} & \textgreater{}0.5 &\cellcolor{gray!30} 0.16 & \cellcolor{gray!30} 0.32 & \cellcolor{gray!30}0.45 & \cellcolor{gray!30}0.25 & \cellcolor{gray!30}0.25 & \cellcolor{gray!30}0.25 & \cellcolor{gray!30}0.35 & \cellcolor{gray!30}0.45 & \cellcolor{gray!30}0.35 & \cellcolor{gray!30}0.35 \\ \hline
 & \textless{}0.5 & 0.32 & 0.43 & 0.5 & 0.25 & 0.3 & 0.35 & 0.45 & 0.45 & 0.4 & 0.45 \\
\multirow{-2}{*}{1} & \textgreater{}0.5 & \cellcolor{gray!30}0.16 & \cellcolor{gray!30} 0.32 & \cellcolor{gray!30}0.4 & \cellcolor{gray!30}0.2 & \cellcolor{gray!30}0.2 & \cellcolor{gray!30}\textless{}0.2 & \cellcolor{gray!30}0.3 & \cellcolor{gray!30}0.35 & \cellcolor{gray!30}0.35 & \cellcolor{gray!30}0.35 \\ \hline
 & \textless{}0.5 & 0.32 & 0.43 & 0.5 & 0.35 & 0.35 & 0.35 & 0.5 & 0.45 & 0.45 & 0.45 \\
\multirow{-2}{*}{2} & \textgreater{}0.5 & \cellcolor{gray!30}0.16 & \cellcolor{gray!30} 0.32 & \cellcolor{gray!30}0.45 & \cellcolor{gray!30}0.25 & \cellcolor{gray!30}0.25 & \cellcolor{gray!30}\textless{}0.2 & \cellcolor{gray!30}0.3 & \cellcolor{gray!30}0.35 & \cellcolor{gray!30}0.35 & \cellcolor{gray!30}0.35
\end{tabular}
}
\caption{Maximum safe end effector velocities --- KUKA iiwa [m/s]. Permissible velocities provided by TS 15066 and mod. TS 15066 taking skin compliance into account (see Eq.~\ref{eq:v_pfl_soft}), and safe velocities determined from empirical measurements that do not exceed the collision force (280~N for $T< 0.5~s$, 140~N for $T > 0.5~s$) `No' --- no skin; `Pas' --- passive skin; `Stop0' / `Stop1' / `Stop1 op' --- active skin and its connection to robot safety inputs. Values in gray are minima of the corresponding rows for first and second phase after impact.}
\label{tab:kuka_isots_comp}
\end{table}

In summary, our results demonstrate the following. For industrial robots (KUKA Cybertech) the effect of active protective skin is rather predictable. It allows for such a robot to detect collisions and respond, thus making human-robot collaboration possible. The passive properties of the protective skin further decrease the impact forces. For collaborative robots, which have their own means of collision detection and response, the situation is more complicated and a number of other factors need to be considered. In this work, we studied the effect of robot velocity, safety settings, and position and impact direction in the workspace. While the effect of velocity is expected, the robot settings---where the skin and possibly other safety sensors are connected---were found to have important and sometimes intriguing effects. In addition, even the different AIRSKIN types, UR-skin vs. Pad, showed different effects on the impact forces. Thus, we conclude that empirical \textit{in situ} measurements are needed and in particular, one should carefully consider the robot, its collision responses (force evolution), and the safety stops available. To illustrate this potential on the example of the UR10e robot, if collisions are the exception rather than the norm in an application and one connects the active skin to the robot E-stop, an almost four times higher velocity (0.5~m/s) will be safe, in contrast to 0.13~m/s that TS~15066~\cite{ISOTS15066} would prescribe.

With collaborative applications becoming more widespread, a better understanding of the actual hazards involved is needed. The relevant standards that application developers have to rely on (ISO/TS 15066~\cite{ISOTS15066}, which will be subsumed by the upcoming standard ISO 10218 \cite{ISO10218}) uses rather coarse and conservative estimates, especially regarding transient contacts. In particular, the standard does not address the effect of soft covers in detail --- it merely suggests to generally use padding or cushioning to reduce the effect of impacts, but fails to back that up with a theory or actual data. This article is a step in extending the theory and providing experimental data on the safety-relevant effects of active and passive soft skins for use in industrial settings. It also shows how active soft skins enable standard industrial robots to be used safely in collaborative applications. This opens up many new applications for robots with higher payloads and reach.

\section*{Acknowledgment}
We thank Bedrich Himmel for assistance with the setup for transient collision measurement. 

\section{Funding}
This work was supported by the Czech Science Foundation (GA CR), project EXPRO (no. 20-24186X). P.S. was additionally supported by the Ministry of Industry and Trade of the Czech Republic (project no. CZ.01.1.02/0.0/0.0/19\_264/0019867).

\bibliographystyle{elsarticle-num}
\bibliography{airskin}

\end{document}